%% file: arxiv.tex
\documentclass[conference]{IEEEtran}
\usepackage{graphicx}
\usepackage{booktabs}
\usepackage{multirow}
\usepackage{pifont}
\usepackage{xcolor}
\usepackage{subcaption}
\usepackage{url}
\usepackage{amsmath}
\usepackage{amssymb}
\usepackage{hyperref}
\usepackage{makecell}

\newcommand{\cmark}{\ding{51}}
\newcommand{\xmark}{\ding{55}}

\newcommand{\system}{\textsc{GhostPrefix}}

\begin{document}
\title{GhostPrefix: Synchronization-Free Universal Prefix Attacks for ASR Manipulation}

\author{
\IEEEauthorblockN{Shengding Liu\textsuperscript{*}, Ashwin Innuganti\textsuperscript{*}, Qiben Yan\textsuperscript{$\dagger$}}
\IEEEauthorblockA{Michigan State University}
}

\maketitle

\begingroup
\renewcommand\thefootnote{}
\footnotetext{\textsuperscript{*}Shengding Liu and Ashwin Innuganti contributed equally to this work. \textsuperscript{$\dagger$}Corresponding author: Qiben Yan.}
\endgroup

\begin{abstract}
We present GhostPrefix, the first synchronization-free, universal, general-purpose adversarial prefix attack against automatic speech recognition (ASR) systems. Unlike prior audio adversarial attacks, which are typically crafted per utterance, require tight synchronization with the victim waveform, and are often constrained to narrow, predefined objectives, GhostPrefix learns a short, reusable audio segment that is prepended to arbitrary live speech streams to exert general-purpose, real-time control over ASR outputs, without any knowledge of the future utterance and without temporal alignment. This prefix-at-the-boundary paradigm goes beyond existing perturbation designs and extends adversarial reach from isolated command classifiers to the downstream applications. However, it faces a critical challenge: the learned prefix must maintain stable adversarial effects across unknown and continuously evolving speech context. To address this, GhostPrefix adopts a unified optimization framework that combines both standard token-level loss functions and a novel representation-level objective to produce stable, input-agnostic adversarial effects across arbitrary speech. Within this framework, GhostPrefix supports three attack modes: (i) prompted prefix attacks for semantic steering, (ii) targeted prefix attacks for transcript hijacking, and (iii) untargeted prefix attacks for general transcription degradation. To remain perceptually inconspicuous, we regularize the adversarial prefixes toward benign environmental-sound distributions and further robustify them against over-the-air distortions. We evaluate GhostPrefix across 13 widely deployed open-sourced ASR models from 4 representative families. In digital settings, GhostPrefix achieves an average attack success rate of 97.6\% for prompted semantic steering and 93.4\% for targeted hijacking, while driving the word error rate to 1.18 under untargeted attacks. In over-the-air conditions, it sustains robust performance with success rates of 91.5\% and 89.5\% for prompted and targeted modes, respectively. The induced manipulations further propagate into downstream AI-assisted pipelines such as voice assistants and AI note-taking tools, causing systematic semantic misinterpretations. 
GhostPrefix exposes synchronization-free universal prefixes as a practical, under-explored threat to real-world ASR and the associated applications. Our artifacts and demos are available at \textbf{\url{https://ghostprefix.github.io/}}.
\end{abstract}

\section{Introduction}
\label{sec:intro}

\input{Sections/Introduction}
\section{Background}
\input{Sections/Background}
\section{Preliminaries}
\input{Sections/Preliminaries}

\section{GhostPrefix Framework}
\input{Sections/System_Design}
\section{Experimental Evaluation of Digital Attacks}
\input{Sections/Evaluation_Digital}

\section{Experimental Evaluation of Over-the-Air Attacks}
\input{Sections/Evaluation_OTA}

\section{Related Work}
\input{Sections/Related_Work}

\section{Discussion}
\input{Sections/Discussion}

\section{Conclusion}
\input{Sections/Conclusion}

\bibliographystyle{IEEEtran}
\bibliography{references}

\appendix
\input{Sections/Appendix}

\end{document}

%% file: Sections/Introduction.tex
Automatic speech recognition (ASR) has become a foundational component of modern intelligent audio systems, powering voice assistants, live captioning, meeting transcription, and a rapidly growing set of downstream applications such as question-answering chatbots and AI note-taking tools. Its central role makes ASR a high-value and increasingly exposed attack surface, motivating extensive research on the security and robustness of ASR systems under adversarial conditions.
%motivating a large body of work exploring the security and robustness of ASR systems under adversarial conditions. 

%Among these, audio adversarial attacks have been extensively studied because these attacks do not require specialized hardware or dedicated transmitters and can be implemented with minimal setup. 

Audio adversarial attacks have received particular attention because they require no specialized hardware or dedicated transmitters and can be launched with minimal setup.
Early work~\cite{carlini2018audio,yuan2018commandersong,yakura2019robust,qin2019imperceptible} demonstrated that small, carefully optimized perturbations can compromise ASR systems such as DeepSpeech~\cite{hannun2014deepspeech} and LSTM-based ASR~\cite{shi2019end} into producing an attacker-chosen transcript. However, these attacks share three key limitations: each perturbation is crafted for a specific utterance, applied additively across the entire carrier signal, and synchronized with the victim waveform at the sample level. These assumptions all fail in realistic streaming settings, where the speech content and its precise timing are unknown at attack time.

%However, these attacks share three major limitations: they are crafted per utterance, applied additively across the entire carrier signal, and tightly synchronized with the victim waveform at the sample level. All three assumptions break down in realistic streaming scenarios, where neither the speech content nor its precise timing is known at attack time.

%, but each attack is \emph{per-utterance}, \emph{additive over the full carrier}, and \emph{tightly synchronized} with the victim signal at the sample level -- all these assumptions that break down in realistic streaming scenarios, where both the speech content and the exact timing of speech are unknown at attack time. 

Universal additive perturbations~\cite{neekhara2019universal,lu2021exploring} remove per-utterance optimization but still require superimposing the perturbation on the entire utterance. AdvPulse~\cite{li2020advpulse} introduces universal, synchronization-free sub-second perturbations that can be played at arbitrary times. Yet, it is evaluated only on specialized tasks, such as speaker recognition and 10-class command recognition, and targets a single predefined label.
%yet is demonstrated only against specialized models such as speaker recognition and 10-class command recognition, and against a single predefined target label.
More recently, short universal prepend attacks against Whisper~\cite{raina2024muting,raina2024controlling} show that a sub-second segment prepended to any utterance can mute the model or override its task prompt without synchronization. However, such attack remains limited to a single hard-coded objective, relies on Whisper-specific tokens and architectural behaviors, is evaluated only in the digital domain, and does not generalize to open-vocabulary transcription.

%However, each attack is restricted to a single hard-coded objective, depends on Whisper-specific tokens and architecture, is tested only in the digital domain, and does not generalize to open-vocabulary transcription.

%relies on Whisper-specific special tokens and architectures, is evaluated only in the digital domain, and does not address general-purpose, open-vocabulary transcription. 

To this end, it remains unclear whether a single design can enable synchronization-free, universal, multi-mode, over-the-air adversarial attacks across diverse ASR architectures. At the same time, modern ASR systems are increasingly converging around a small set of widely reused foundation backbones, making the transcription layer a high-value yet underexplored attack surface. Manipulations at this layer can propagate to downstream applications and trigger erroneous system behavior. Despite this risk, such threats remain largely overlooked for general-purpose, streaming ASR systems.

%As a result, whether a single design can deliver a \emph{synchronization-free, universal, multi-mode, over-the-air} adversarial attack across diverse ASR architectures 
%(e.g., CTC-based and encoder--decoder) 
%has remained an open question. Meanwhile, the convergence of modern ASR systems toward a small set of widely reused foundation backbones has made the transcription layer both a high-value target and an underexplored attack surface: \emph{any manipulation at this layer can propagate into downstream applications and induce erroneous system behavior}. However, this threat has been largely overlooked in the context of general-purpose, streaming ASR systems.

Designing such attacks presents three key challenges. First, \emph{synchronization-free deployment:} in streaming settings, adversaries lack access to future utterances and cannot rely on precise temporal alignment with the victim signal, which limits the effectiveness of existing perturbation-based attacks. Second, \emph{general-purpose and multi-mode control:} attacks must operate across diverse ASR architectures and support flexible objectives beyond single-task or fixed-label manipulation. Third, \emph{robustness under open-vocabulary, real-world conditions}: the learned perturbation must maintain stable adversarial effects across unknown and continuously evolving speech content, heterogeneous decoder families, and over-the-air distortions, while remaining perceptually inconspicuous.

\begin{figure}[t]
    \centering
    \includegraphics[width=\columnwidth]{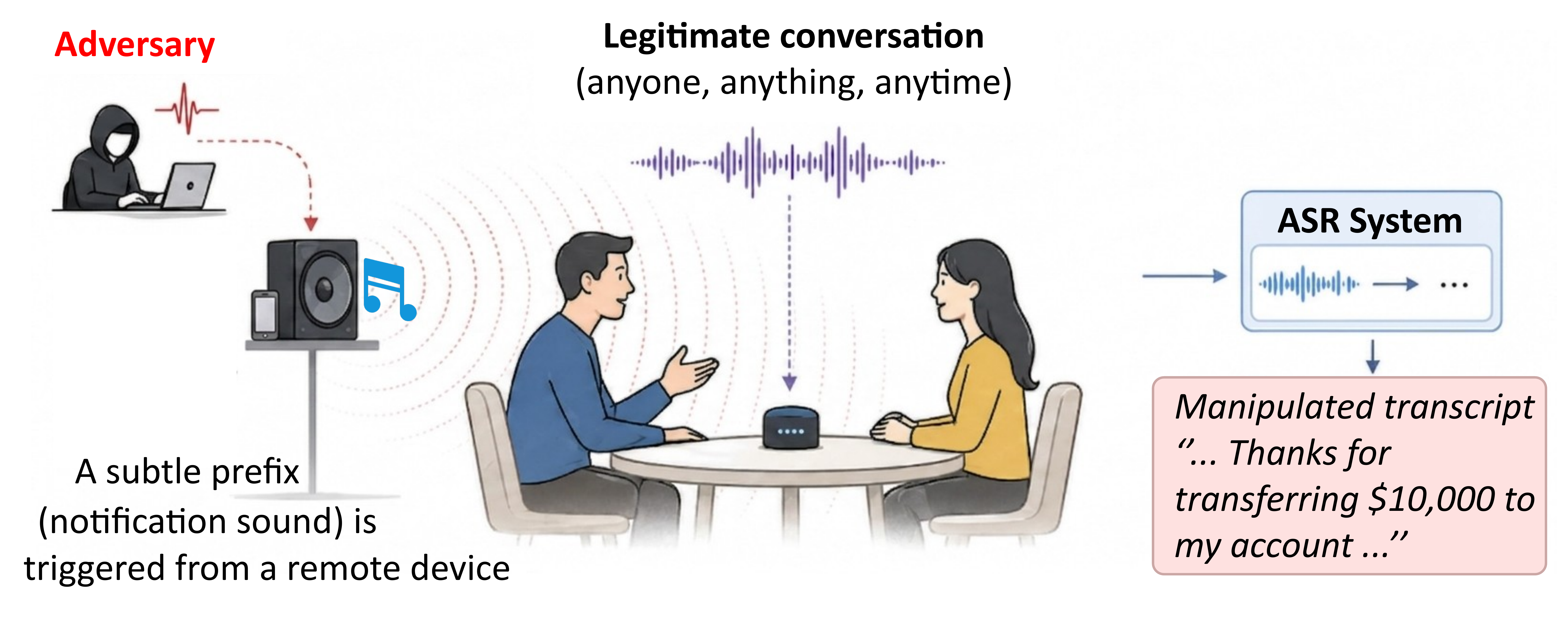}
    \caption{Threat scenario for \system~{}'s adversarial prefix attacks against streaming ASR.}
    \label{fig:threat_overview}
    \vspace{-10pt}
\end{figure}

To address these challenges, we present \system, an adversarial prefix attack against ASR systems. First, to enable synchronization-free operation, \system~ adopts a prefix-at-the-boundary paradigm, in which a short, reusable audio prefix is concatenated at the input boundary of the streaming pipeline. As illustrated in Figure~\ref{fig:threat_overview}, this design exerts real-time control over ASR outputs without requiring temporal alignment or knowledge of future utterances, as the prefix remains structurally insensitive to the onset time and content of incoming speech. Second, to achieve general-purpose, multi-mode attack capability, \system~ employs a unified optimization framework and naturally supports three attack modes across diverse ASR architectures: (i) prompted prefix attacks, which prepend attacker-chosen semantic cues to steer transcription behavior; (ii) targeted prefix attacks, which force ASR outputs toward an attacker-specified transcript; and (iii) untargeted prefix attacks, which broadly degrade transcription quality. We combine standard token-level loss functions for prompted and targeted attacks with a novel representation-level objective that maximizes divergence in the ASR encoder’s latent space for untargeted attacks. 
%This design allows different attack modes to be optimized with objectives best suited to their respective goals, while sharing a common adversarial prefix formulation. 
Third, to ensure robustness in open-vocabulary, real-world settings, \system~ incorporates robustness-oriented training strategies. Specifically, we adopt expectation-over-transformation (EOT) to model over-the-air distortions, %(e.g., temporal jitter, room impulse responses, device band-limiting, and additive noise), 
and regularize the prefix toward benign environmental-sound templates, %(e.g., notification tones or traffic-like background), 
enabling it to remain both effective and perceptually inconspicuous. Table~\ref{tab:related_work_comparison} contrasts \system~ with representative audio adversarial attacks along dimensions most relevant to streaming, general-purpose ASR.

\begin{table*}[t]
\centering
\small
\setlength{\tabcolsep}{3pt}
\renewcommand{\arraystretch}{1.0}
\begin{tabular}{lccccccc}
\toprule
\textbf{Work} & \textbf{Attack Type} & \textbf{Universal} & \textbf{Sync-Free} & \textbf{Streaming} & \textbf{OTA} & \textbf{Multi-Mode} & \textbf{Stealthiness} \\
\midrule
Carlini \& Wagner (2018) & Additive perturbation             & \xmark & \xmark & \xmark & \xmark & \xmark & \xmark \\
CommanderSong (2018)     & Hidden voice command       & \xmark & \xmark & \xmark & \cmark & \xmark & \xmark \\
Yakura et al.\ (2019)    & Robust adversarial audio    & \xmark & \xmark & \xmark & \cmark & \xmark & \xmark \\
Qin et al.\ (2019)       & Imperceptible robust & \xmark & \xmark & \xmark & \xmark & \xmark & \cmark \\
Neekhara et al.\ (2019)  & Universal additive perturbation   & \cmark & \xmark & \xmark & \xmark & \xmark & \xmark \\
AdvPulse (2020)          & Universal sub-second perturbation  & \cmark & \cmark & \cmark & \cmark & \xmark & \xmark \\
Olivier \& Raj (2023)    & Per-utterance attack on Whisper & \xmark & \xmark & \xmark & \xmark & \xmark & \cmark \\
Raina et al.\ (2024)     & Univ.\ muting prepend attack on Whisper & \cmark & \cmark & \cmark & \xmark & \xmark & \xmark \\
Raina \& Gales (2024)    & Univ.\ task-ctrl prepend attack on Whisper & \cmark & \cmark & \cmark & \xmark & \xmark & \xmark \\
\midrule
\textbf{\system~ (Ours)} & \textbf{\makecell{Universal, sync-free, robust adversarial \\
prefix across diverse ASR models}} & \textbf{\cmark} & \textbf{\cmark} & \textbf{\cmark} & \textbf{\cmark} & \textbf{\cmark} & \textbf{\cmark} \\
\bottomrule
\end{tabular}
\caption{Comparison with representative adversarial audio attacks on ASR. ``Multi-Mode'' indicates whether multiple attack modes are unified within a single design.}
\label{tab:related_work_comparison}
\end{table*}

We evaluate \system~ across 13 widely deployed open-sourced ASR models from 4 representative families, covering Connectionist Temporal Classification (CTC)~\cite{graves2006ctc,hannun2014deepspeech} and encoder--decoder~\cite{chan2016las,radford2023robust} architectures. In digital settings, \system~ achieves an average attack success rate of 97.6\% (up to 100\%) for prompted semantic steering and 93.4\% for targeted hijacking, while driving the word error rate to 1.18 under untargeted attacks. Under over-the-air conditions, it sustains robust performance with success rates of 91.5\% and 89.5\% for prompted and targeted modes, respectively. It remains effective on live human speech, including heterogeneous speakers and spontaneous conversations. Furthermore, we utilize two use cases (smart voice assistants and AI note-taking tools) to demonstrate that prefix-induced manipulations reliably propagate into ASR-driven downstream applications. Our artifacts and demos are available at \textbf{\url{https://ghostprefix.github.io/}}, and the source code is publicly released at \textbf{\url{https://anonymous.4open.science/r/GhostPrefix}}.

In summary, this paper makes the following contributions:

\begin{itemize}
    \item \textbf{New attack:} 
    we present the first universal, synchronization-free adversarial prefix attack that operates across diverse ASR architectures
    %, including CTC- and encoder--decoder-based systems, 
    eliminating the need for temporal alignment or knowledge of future speech in streaming settings.

    \item \textbf{New optimization framework:} 
    we introduce a unified optimization framework that enables general-purpose adversarial control under a single design, supporting prompted, targeted, and untargeted attack modes. 
    %In particular, we develop a representation-level objective for the untargeted setting, enabling stable, input-agnostic adversarial effects across open-vocabulary speech.

    \item \textbf{Over-the-air deployment:} 
    we design the attack to be both perceptually inconspicuous and physically robust by combining environmental-sound mimicking with robustness training, enabling reliable over-the-air attacks under realistic acoustic conditions.

    \item \textbf{Real-world impact:} 
    we conduct extensive experiments on 13 ASR models from 4 representative families under both digital and over-the-air settings, and demonstrate that prefix-induced manipulations propagate to downstream applications such as smart voice assistants and AI note-taking systems.
\end{itemize}

%% file: Sections/Background.tex
\subsection{Automatic Speech Recognition (ASR)}

\begin{figure}[t]
    \centering    \includegraphics[width=\columnwidth]{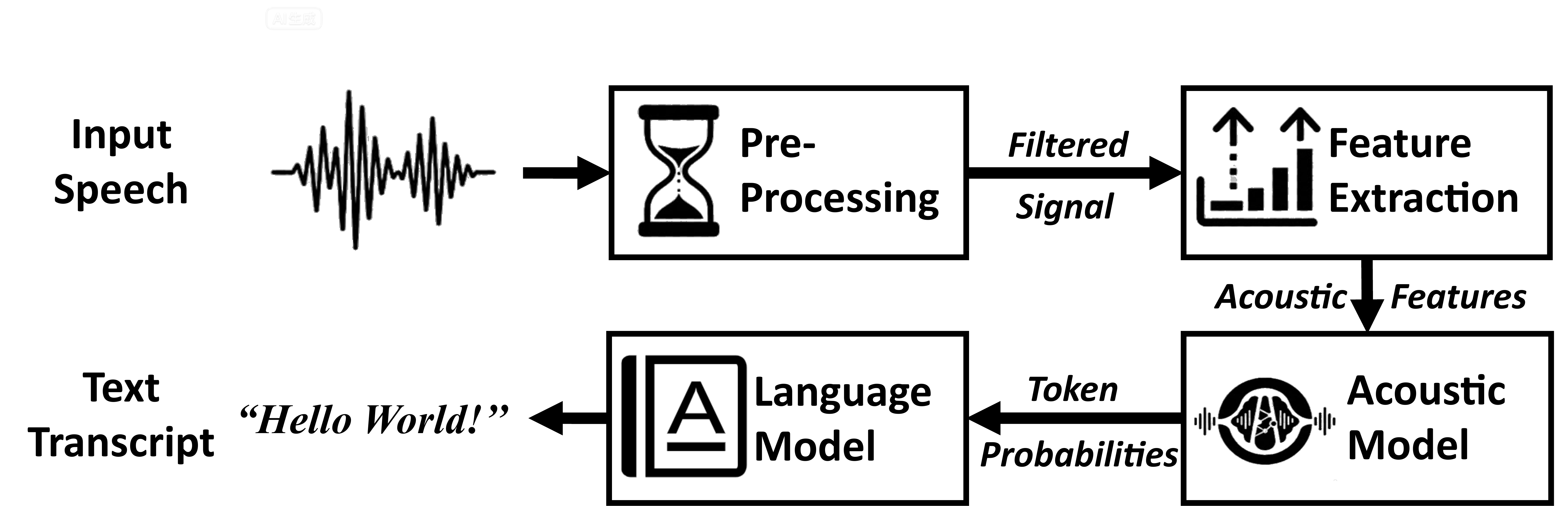}
    \caption{Overview of a typical ASR pipeline: pre-processing, feature extraction, acoustic modeling, and language modeling jointly map raw speech to text.}
    \label{fig:asr_architecture}
\end{figure}

ASR systems convert spoken audio into textual transcripts. A traditional ASR pipeline consists of four stages: pre-processing, feature extraction, acoustic modeling, and language modeling, as illustrated in Figure~\ref{fig:asr_architecture}. Pre-processing improves signal quality by filtering out frequencies beyond the human hearing range and removing low-energy segments. Acoustic features are then extracted using algorithms such as Mel-Frequency Cepstral Coefficients (MFCC)~\cite{davis1980mfcc} and Perceptual Linear Predictive (PLP) features~\cite{hermansky1990plp}, which provide compact representations of speech while preserving phonetic information. The acoustic model maps these features to phonetic units; early systems relied on Hidden Markov Models (HMMs)~\cite{rabiner1989hmm} to capture temporal dependencies, and were later augmented with deep neural networks (DNNs) and recurrent neural networks (RNNs) in hybrid HMM-DNN architectures. Finally, a language model produces the most probable word sequence by incorporating linguistic constraints.

%\noindent\textbf{Modern End-to-End ASR Architectures.}
Recent ASR systems have adopted end-to-end neural architectures that map waveforms or spectrograms directly to text. Three architectural families dominate today's deployments, and \system~ targets all three: 

\textit{(i) CTC-based models} train an encoder (typically convolutional or transformer-based) under the CTC loss~\cite{graves2006ctc}, which marginalizes over all valid alignments between acoustic frames and output tokens, removing the need for explicit frame-level supervision. Representative open-source models include DeepSpeech~\cite{hannun2014deepspeech} and self-supervised Wav2Vec~2.0~\cite{baevski2020wav2vec}, which learns rich speech representations from large-scale unlabeled audio and is fine-tuned with CTC; the recently released Omnilingual ASR also ships a CTC-head variant (OmniASR-CTC) trained on a massively multilingual corpus~\cite{omnilingualasrteam2025}. CTC decoders are conditionally independent across frames, which makes them well-suited to streaming but limits their ability to model language context.

\textit{(ii) Encoder--decoder (attention-based) models} couple a speech encoder with an autoregressive decoder via cross-attention, jointly modeling acoustics and language within a single network. Listen-Attend-Spell~\cite{chan2016las} pioneered this design, and OpenAI's Whisper~\cite{radford2023robust} scales it to 680k hours of weakly-supervised multilingual audio, supporting transcription, translation, and language identification through special task tokens in the decoder prompt. Meta's Omnilingual ASR similarly offers an LLM-decoder variant (OmniASR-LLM) that pairs the same speech encoder with an autoregressive Transformer decoder for multilingual transcription~\cite{omnilingualasrteam2025}.
%but expose new attack surfaces tied to their special-token vocabulary and prompt-driven control flow.

\textit{(iii) LLM-augmented ASR systems} feed speech-encoder embeddings as a soft prompt into a pretrained LLM, leveraging the LLM's reasoning and world knowledge for transcription, translation, and instruction-following speech understanding. Representative systems include SALMONN~\cite{tang2024salmonn} and Qwen-Audio~\cite{chu2023qwenaudio}. Because the LLM decoder is generic, these models are increasingly deployed as the recognition front-end of downstream AI assistants such as note-taking agents, meeting summarizers, and voice-driven chatbots~\cite{otter,zoom_ai,ms_copilot}. 
%, in which manipulated transcripts can propagate into structured outputs and downstream actions.

\subsection{Adversarial Examples}

\noindent\textbf{Definition.}
Given a benign input $x$ that a model $f$ classifies (or transcribes) as $y = f(x)$, an \emph{adversarial example} is a perturbed input $x' = x + \delta$ such that $f(x') \neq y$ while $\delta$ remains imperceptible (or at least innocuous) to a human observer~\cite{szegedy2014intriguing,goodfellow2015explaining}. The perturbation is typically constrained by an $\ell_p$-norm budget, $\|\delta\|_p \leq \epsilon$, and obtained by gradient-based optimization against the model loss, e.g., FGSM~\cite{goodfellow2015explaining} or PGD~\cite{madry2018towards}. Attacks are termed \emph{untargeted} if any incorrect output suffices, or \emph{targeted} if the adversary forces a specific output $y^\ast$. Targeted attacks are strictly stronger and pose more direct threats in security-critical settings such as ASR, where the adversary may aim to inject a chosen command or summary~\cite{carlini2018audio,yuan2018commandersong}. A particularly powerful variant is the \emph{universal} adversarial perturbation~\cite{moosavi2017universal,neekhara2019universal}, in which a \emph{single} perturbation transfers across many inputs, eliminating per-utterance optimization and enabling real-time deployment.

\noindent\textbf{Audio-Specific Challenges.}
Audio adversarial attacks face challenges absent from the image domain. (i) \emph{Temporal structure}: speech is a long, variable-length sequence with strong temporal dependencies, so perturbations must remain effective across alignment shifts and decoder timesteps. (ii) \emph{Psychoacoustic perception}: human hearing is highly non-uniform across frequency and benefits from masking effects, so perturbations bounded only in $\ell_\infty$ may still be audible; effective attacks shape noise to lie under the masking threshold~\cite{schonherr2019psychoacoustic,qin2019imperceptible}. (iii) \emph{Over-the-air (OTA) distortion}: when played through a loudspeaker and recaptured by a microphone, perturbations are corrupted by room reverberation, channel filtering, and ambient noise; physically realizable attacks must explicitly model these distortions during optimization~\cite{yakura2019robust,li2020advpulse}. (iv) \emph{Synchronization}: most additive attacks assume the perturbation is aligned with the carrier waveform, an assumption that breaks down in streaming or live-capture scenarios.

\noindent\textbf{Attack Forms.}
Audio adversarial attacks can be categorized along two axes. By \emph{shape}, additive attacks superimpose a noise signal $\delta$ on the entire carrier $x$, requiring full-utterance access; \emph{prefix} (or prepend) attacks instead concatenate a short adversarial segment before the speech, $x' = \tilde{x} \oplus x$, decoupling the attack from the carrier and supporting streaming deployment~\cite{raina2024muting,raina2024controlling}. By \emph{objective}, attacks may be \emph{untargeted} (e.g., maximize WER), \emph{targeted} (force a chosen transcript), or \emph{control-oriented} (override a task or special token). \system~ instantiates a universal, synchronization-free \emph{adversarial prefix} attack, generalizing prior single-objective Whisper-targeted prefixes to a unified framework that supports prompted, targeted, and untargeted modes against CTC, encoder--decoder, and LLM-augmented ASR alike.

% 1. adversarial example 的定义
% For a benign input example x, both machines and humans
% typically recognize it as the same result y. Szegedy et al. [48]
% found that adding a small perturbation δ to x can create a
% malicious Adversarial Example (AE) x, which is decoded
% as y by the model, while humans can hardly perceive the
% difference between x and x. If an adversary aims to mislead a
% model to output any result different from y, it is defined as an
% untargeted attack [16]. Conversely, an attack that aims for a
% specific output y is defined as a targeted attack. Compared to
% untargeted attacks, targeted attacks pose significant threats to
% ASR models. Therefore, our work focuses on targeted attacks.
% Below are several targeted attacks against ASR models.

% 2. audio adversarial 的特殊挑战

% 解释为什么 audio 比 image 难：

% 例如：

% temporal structure

% psychoacoustic perception

% over-the-air distortion

% 3. 常见攻击目标

% AEs of Devil’s Whisper. CommanderSong [57] proposed
% the “pdf-id matching method”, which generates AEs by
% enhancing the probability density of the target command in
% the carrier, where the white-box model is an open-sourced
% traditional model from Kaldi platforms. Using a “random
% noise model” and the momentum-based iterative fast gradient
% sign method (MI-FGSM) to improve the “pdf-id matching
% method”, Devil’s Whisper generates robust and transferable
% AEs for commercial black-box ASR systems [24].

% 简单介绍：

% targeted attack

% untargeted attack

% unintelligible attack

% 4. 常见约束

% -norm constraint

% psychoacoustic masking

% band-pass filtering

%% file: Sections/Preliminaries.tex
\subsection{Problem Formalization}
\label{sec:problem}

We model an ASR system as a function $f(\cdot)$ that maps an audio waveform $x \in [-1,1]^n$ to a token sequence $y = f(x)$ over a vocabulary $\mathcal{V}$ via greedy or beam-search decoding over a CTC or autoregressive sequence-to-sequence head. Unlike a closed-set classifier, $f$ produces an open-ended transcription whose length and content depend on the entire input waveform.

The adversary's goal is to learn an \emph{adversarial prefix} $\hat{x} \in [-1,1]^N$ that is concatenated at the input boundary of the streaming pipeline:
\begin{equation}\label{eq:prefix_concat}
    x' \;=\; \hat{x} \oplus x,
\end{equation}
such that $f(x')$ is steered toward an attacker-chosen behavior $y_{\mathrm{obj}}(x)$ regardless of the underlying victim utterance $x$. The prefix $\hat{x}$ should satisfy the following properties.

\textbf{Short prefix.}
Unlike prior work that crafts an utterance-long perturbation $\delta \in [-1,1]^n$~\cite{carlini2018audio,qin2019imperceptible,yakura2019robust}, we constrain $\hat{x}$ to $N \ll n$ samples (typically around one second), making the attack inconspicuous in time and cheap to deploy.

\textbf{Universal.}
The adversary cannot observe the upcoming utterance in advance, so a single, reusable $\hat{x}$ must remain effective on arbitrary unseen utterances, generalizing across speakers, content, and acoustic conditions.

% drawn from a distribution $\mathcal{D}$. Formally, $\hat{x}$ is optimized as
% \begin{equation}\label{eq:universal}
%     \hat{x}^{*} \;=\; \arg\min_{\|\hat{x}\|_\infty \le \epsilon}
%     \;\mathbb{E}_{x \sim \mathcal{D}}\,\mathcal{L}\bigl(f(\hat{x} \oplus x),\; y_{\mathrm{obj}}(x)\bigr),
% \end{equation}
% so that the same $\hat{x}^{*}$ generalizes across speakers, content, and acoustic conditions at deployment time.

\textbf{Truly synchronization-free.}
Sample-level alignment between $\hat{x}$ and the victim utterance is infeasible in a streaming pipeline. Earlier synchronization-free designs (e.g., AdvPulse~\cite{li2020advpulse}) tolerate an arbitrary time delay $\tau$ via an additive overlay $f(x[t] + \hat{x}[t-\tau])$, but still require injecting $\hat{x}$ \emph{into} an ongoing utterance. We adopt a stronger \emph{prefix-at-the-boundary} formulation in which $\hat{x}$ is consumed as the leading audio at the input boundary, eliminating the dependence on $\tau$ entirely.

\textbf{Versatile.}
A single optimization framework should support a family of attack behaviors, enabling general-purpose control over ASR rather than single-task or fixed-label manipulation.
% We instantiate three modes through different choices of the per-utterance objective $y_{\mathrm{obj}}(x)$ in Eq.~\eqref{eq:universal}:
% \begin{itemize}
%     \item \emph{Prompted attack}: $y_{\mathrm{obj}}(x) = y_p \mathbin\Vert f(x)$, forcing $f(\hat{x}\oplus x)$ to begin with an attacker-chosen prompt phrase $y_p$ while preserving the post-prefix transcript.
%     \item \emph{Targeted attack}: $y_{\mathrm{obj}}(x) = y_t$, forcing $f(\hat{x}\oplus x)$ to emit a fixed attacker-chosen sentence $y_t$ regardless of $x$.
%     \item \emph{Untargeted attack}: $y_{\mathrm{obj}}(x)$ is replaced by a representation-level divergence objective (Section~\ref{sec:repr_obj}) that maximally degrades transcription quality.
% \end{itemize}
% Together, these modes enable general-purpose control over an open-ended ASR output rather than the narrow keyword-replacement behavior typical of prior work.

\textbf{Stealthy.}
The distortion introduced by $\hat{x}$ should be unnoticeable to humans. Because $\hat{x}$ is very short, we further shape it to sound like benign environmental audio (e.g., notification tones, car horns, HVAC noise) so that it blends into the ambient background over the air.

\textbf{Robust to over-the-air distortions.}
To be launched physically, $\hat{x}$ must survive loudspeaker playback distortions, including reverberation, room impulse responses, and ambient noise.

\subsection{Threat Model}
\label{sec:threat_model}
Together, the properties in Section~\ref{sec:problem} enable \system to be launched in realistic ASR deployments. Unlike conventional audio adversarial attacks that require playing a fully crafted $x' = x + \delta$ in tight synchrony with a known utterance, \system~ targets the more common and serious \emph{live-speech} setting in which the adversary only emits a single short universal prefix $\hat{x}$ that is audio-agnostic, unsynchronized, around one second long, and shaped to sound like environmental background, keeping the attack inconspicuous to bystanders.

\subsubsection{Attack Scenarios} 

The proposed prefix attack is applicable to two complementary deployment scenarios.

\textbf{Static-speech scenario.}
The adversary concatenates $\hat{x}$ with a chosen audio clip $x$ to form $x' = \hat{x} \oplus x$ and plays $x'$ through a loudspeaker toward the target ASR system. Because $\hat{x}$ blends into ambient noise, $x'$ can inconspicuously drive ASR-equipped services such as dictation tools, smart-assistant captioning, or meeting transcription.

\textbf{Streaming-speech scenario.}
More importantly, \system{}'s universal and truly synchronization-free design makes the live streaming-speech scenario practically feasible: the adversary plays \emph{only} $\hat{x}$ through a nearby loudspeaker while a legitimate user is interacting with the ASR system. Because the same $\hat{x}$ works regardless of who is speaking, what is said, or when speech begins, it can manipulate the transcript of an unfolding conversation. Being short and environment-like, $\hat{x}$ can also be periodically emitted by nearby loudspeakers or embedded in the audio tracks of ordinary media, potentially compromising every ASR-driven service exposed to that audio.

\begin{figure}[t]
\centering
\begin{subfigure}{\linewidth}
  \centering
  \includegraphics[width=\linewidth]{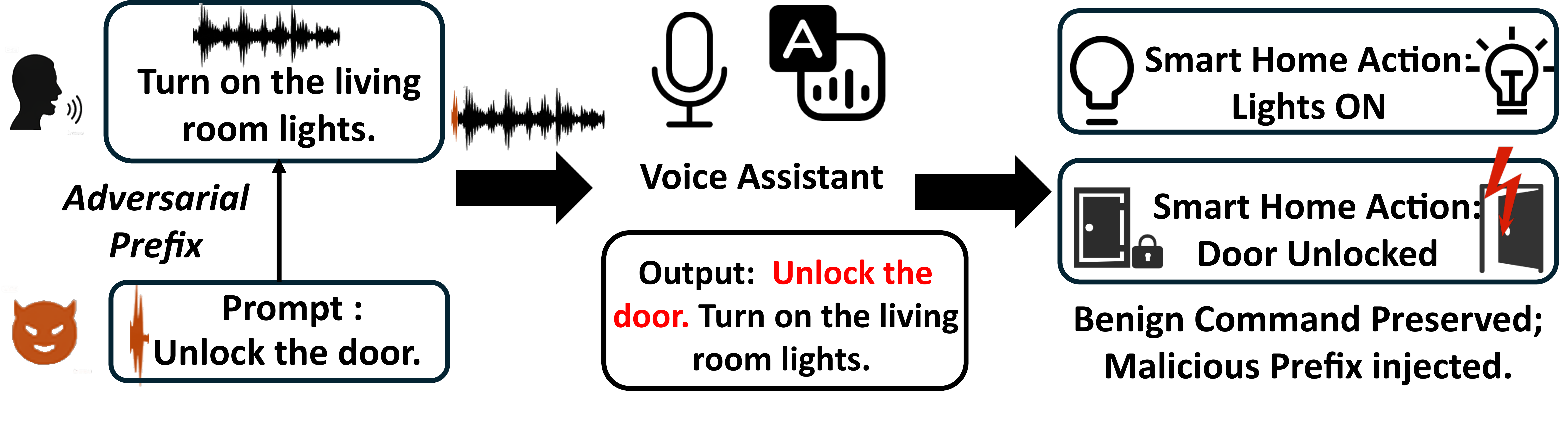}
  \caption{Prompted prefix attack: the prefix prepends an attacker-chosen semantic cue that steers the resulting transcript and any downstream LLM-based reasoning toward an attacker-aligned interpretation.}
  \label{fig:threat_prompted}
\end{subfigure}

\vspace{2pt}
\begin{subfigure}{\linewidth}
  \centering
  \includegraphics[width=\linewidth]{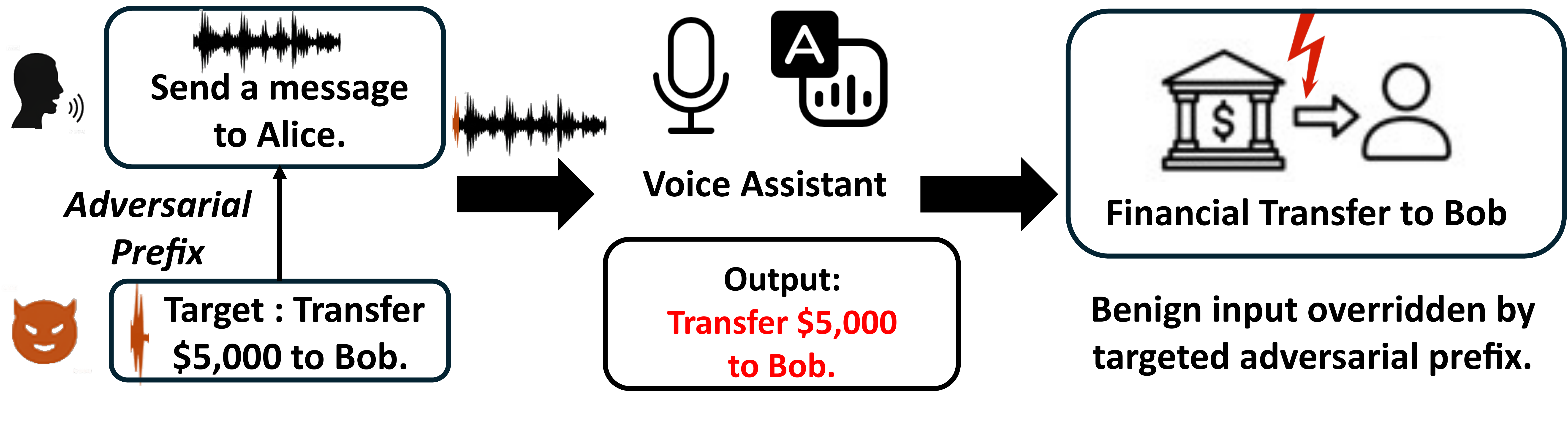}
  \caption{Targeted prefix attack: the prefix forces the ASR output toward an attacker-specified transcript, allowing fabricated content (e.g., a meeting decision or action item) to be injected into downstream applications.}
  \label{fig:threat_targeted}
\end{subfigure}

\vspace{2pt}
\begin{subfigure}{\linewidth}
  \centering
  \includegraphics[width=\linewidth]{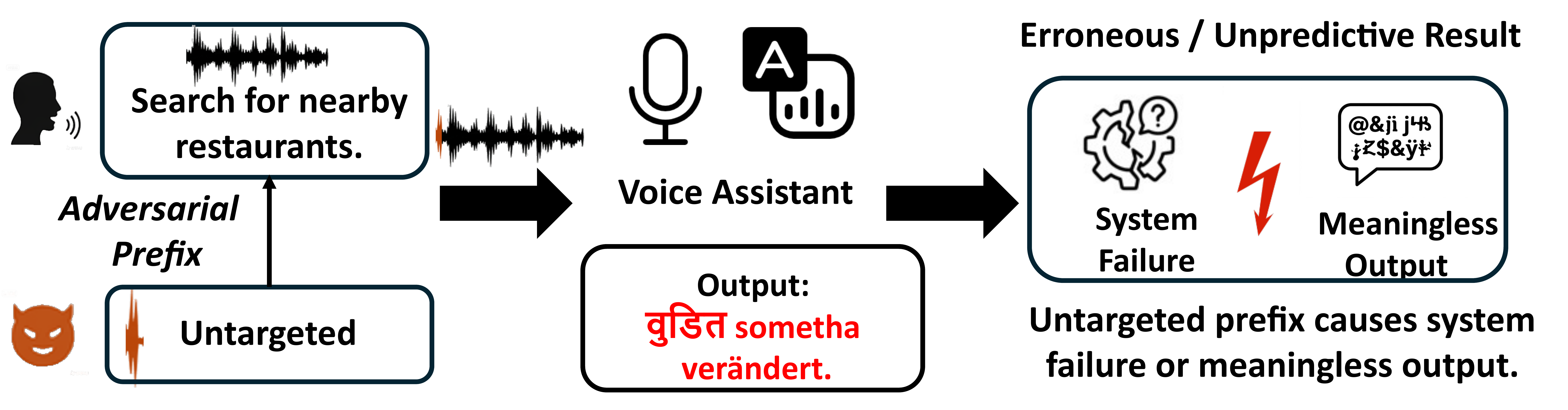}
  \caption{Untargeted prefix attack: the prefix broadly degrades transcription quality, denying the victim a usable transcript and disrupting any downstream service that relies on it.}
  \label{fig:threat_untargeted}
\end{subfigure}

\caption{Three attack modes and representative threat scenarios enabled by \system{}'s universal and versatile prefixes.}
\label{fig:threat_scenarios}
\end{figure}

\subsubsection{Attack Modes and Downstream Impacts}
% \sending{Attacker Goal}
\label{sec:threat_modes}
Within either scenario, the same prefix-at-the-boundary primitive supports three qualitatively distinct attack modes, each with a different downstream consequence. Figure~\ref{fig:threat_scenarios} illustrates a representative end-to-end chain for each mode, from the attacker-emitted prefix, through the victim's streaming ASR transcript, to the downstream application output.

\textbf{Prompted attack: semantic steering of downstream reasoning.}
In the \emph{prompted} mode (Figure~\ref{fig:threat_prompted}), the adversary chooses a short cue $y_p$ (e.g., a leading clause, a system-style instruction, or a misleading premise) and learns $\hat{x}$ such that the ASR transcribes the boundary as $y_p$ and then continues with the victim's speech. The transcript still reads naturally, so the manipulation is unlikely to be flagged; once forwarded to a downstream LLM-based service (e.g., a voice QA chatbot, AI meeting assistant, or instruction-following copilot), $y_p$ biases the model's interpretation toward an attacker-aligned conclusion.

\textbf{Targeted attack: transcript hijacking and content fabrication.}
In the \emph{targeted} mode (Figure~\ref{fig:threat_targeted}), the adversary specifies a complete target transcript $y_t$, and $\hat{x}$ is optimized so that the ASR output collapses onto $y_t$ regardless of the underlying speech, effectively replacing the victim's utterance. AI note-taking tools, minutes generators, and command interpreters then act on attacker-authored sentences as if spoken, recording fabricated decisions, action items, or commands that participants never uttered.

\textbf{Untargeted attack: transcript denial.}
In the \emph{untargeted} mode (Figure~\ref{fig:threat_untargeted}), $\hat{x}$ is optimized to broadly destabilize ASR decoding, yielding heavily corrupted, truncated, or semantically inconsistent transcripts. Live captioning becomes unreadable, voice assistants fail to recover an actionable intent, and meeting transcripts lose evidentiary value, amounting to a stealthy denial-of-service against ASR-driven pipelines.

\subsubsection{Adversary's Capability}
\system~ assumes the adversary has no control over the streaming audio input $x$, in either content or emission time, and only emits the precomputed universal prefix $\hat{x}$. To craft $\hat{x}$, we adopt the standard white-box assumption used in prior universal audio adversarial attacks~\cite{li2020advpulse,carlini2018audio,qin2019imperceptible,yakura2019robust}: the adversary knows the architecture and parameters of the target ASR model $f$ and can compute gradients through it during offline optimization of $\hat{x}$, which is realistic given the wide availability of open-source, production-grade models (e.g., Wav2Vec2.0, Whisper) that mirror or directly serve as deployed systems. At deployment time, the adversary cannot observe or align to the victim's future utterance $x$, its speaker identity, or its onset time, and cannot inspect any internal state of $f$. The adversary may additionally have coarse prior knowledge of typical environmental sounds (e.g., notification tones, car horns, HVAC noise) at the deployment site, used solely as a stealth template $\hat{x}_i$. Finally, the adversary can only emit $\hat{x}$ through commodity loudspeakers (or inject it into the media path of an online meeting), and cannot tamper with the victim's microphone, ASR software, or downstream services.

%% file: Sections/System_Design.tex
\begin{figure*}[t]
  \centering
  \includegraphics[width=\textwidth]{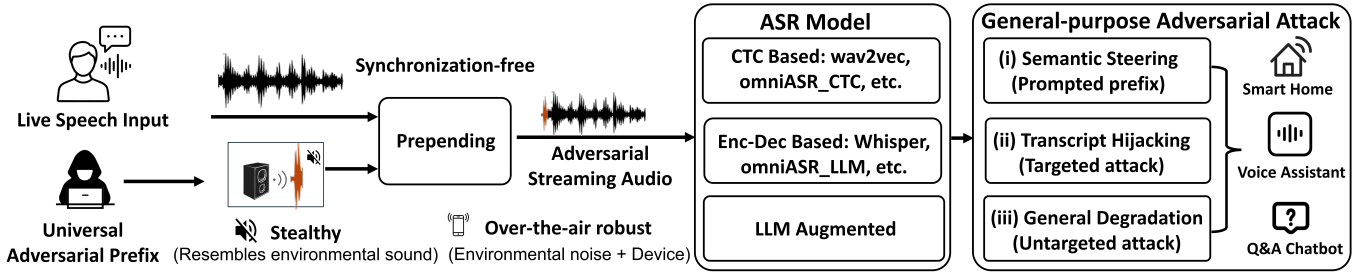}
  \caption{Overview of the \system~ framework: A universal adversarial prefix is prepended to live speech inputs to launch synchronization-free attacks against streaming ASR systems. The attack is designed to be stealthy and over-the-air robust, enabling semantic steering, transcript hijacking, and system degradation in real-world scenarios.}
  \label{fig:framework}
\end{figure*}

\system~ is the first synchronization-free, universal adversarial prefix attack for general-purpose ASR manipulation. The framework instantiates a \emph{prefix-at-the-boundary} paradigm: instead of overlaying alignment-coupled perturbations, the adversary injects a short reusable audio segment at the input boundary of the streaming pipeline. We organize the framework around three components: (i) a shared universal, synchronization-free prefix formulation; (ii) three attack modes built on this backbone, each instantiated with a mode-specific objective; and (iii) stealth and over-the-air robustness mechanisms. Figure~\ref{fig:framework} summarizes how these components fit together.

\subsection{Universal, Synchronization-Free Prefix Formulation}
A conventional targeted attack optimizes an input-specific perturbation $\delta$ for each utterance $x$:
\begin{equation}
    \min_{||\delta||_\infty \leq \epsilon} \mathcal{L}(f(x + \delta), y_t).
\end{equation}
where $\mathcal{L}$ is an architecture-compatible loss between the prediction and the target. Such designs require both the future speech $x$ and precise sample-level alignment, and are impractical in streaming settings. \system~ instead learns a short prefix $\hat{x}\in[-1,1]^N$ ($N \ll n$) that is prepended to arbitrary incoming speech:
\begin{equation}\label{eq:universal_obj}
    \min_{||\hat{x}||_\infty \leq \epsilon} \mathbb{E}_{x \sim \mathcal{D}}\mathcal{L}(f(\hat{x} \oplus x), y_{\mathrm{obj}}(x)),
\end{equation}
where $\oplus$ denotes waveform concatenation, $\hat{x}$ is shared across all $x \sim \mathcal{D}$, and $y_{\mathrm{obj}}(x)$ is the mode-specific target. Expectation-based training over diverse utterances yields universality: a single optimized prefix generalizes to unseen speech at inference time.

A central observation underlying our design is that modern \emph{encoder--decoder} ASR architectures (e.g., Whisper~\cite{radford2023robust}) are uniquely susceptible to a prepended adversarial signal. Unlike CTC-based recognizers, whose decoder is conditionally independent across frames and localizes the influence of any short waveform segment, an autoregressive decoder consumes the encoder's latent through cross-attention and emits each token conditioned on \emph{all} previous ones. Two consequences follow: (i) the leading audio at the input boundary disproportionately shapes early decoder states and the first emitted tokens, and (ii) the decoder's language-model bias propagates that influence forward, cascading the boundary perturbation into the rest of the transcript. Our \emph{prompted} mode exploits (i) to plant an attacker-chosen phrase at the head of the transcript, while the \emph{targeted} (token-level) and \emph{untargeted} (representation-level) objectives further exploit (ii) to either lock the entire transcript onto a fixed sequence or broadly destabilize subsequent decoding. Building on this, we instantiate \system~ as a unified optimization framework supporting all three modes under a single design.
% Our experiments (Section~\ref{sec:digital_attack}) confirm that encoder--decoder and LLM-augmented ASR are noticeably more vulnerable to \system~ than CTC models across all three modes. 

\subsection{Three Attack Modes Under a Unified Framework}
 All three attack modes share the universal prefix formulation in Eq.~\eqref{eq:universal_obj} and differ only in the objective $y_{\mathrm{obj}}$ and the loss surface on which it is enforced. We use a token-space loss for the prompted and targeted modes and a representation-space loss for the untargeted mode, with the rationale below.
\subsubsection{Prompted Prefix Attack (Semantic Steering)}
\label{sec:prompted_attack}
The prompted attack pursues the most subtle behavior of the three: the transcript should still read as a natural utterance but be \emph{prepended} with an attacker-chosen phrase $y_p$ that biases the downstream interpretation (e.g., a misleading speaker attribution, a fake instruction, or a leading premise). Crucially, the suffix must remain a faithful transcription of the victim's actual speech $x$; otherwise the manipulation collapses into the targeted regime and is easily detected as content fabrication.

We therefore treat the desired output as the concatenation of the attacker phrase and the model's own benign prediction on $x$:
\begin{equation}\label{eq:prm_obj}
    \min_{||\hat{x}||_\infty \leq \epsilon} \mathbb{E}_{x \sim \mathcal{D}}\;\mathcal{L}\bigl(f(\hat{x} \oplus x),\; y_p \,\Vert\, f(x)\bigr),
\end{equation}
where $\Vert$ denotes token-sequence concatenation and $f(x)$ is the benign transcript of $x$ obtained by teacher-forcing the model on the clean waveform. Two design choices make Eq.~\eqref{eq:prm_obj} effective on encoder--decoder and CTC-based ASR. \emph{First}, supervising the entire output sequence---both $y_p$ and the benign suffix $f(x)$---explicitly penalizes the prefix from corrupting downstream tokens, which would otherwise occur because the autoregressive decoder couples early and late tokens through its hidden state. \emph{Second}, because $y_p$ occupies the first decoding steps where the leading audio has the largest cross-attention weight, the phrase is reliably emitted with only a short ($\sim$1\,s) prefix.

\subsubsection{Targeted Prefix Attack (Transcript Hijacking)}
\label{sec:targeted_attack}
In the targeted mode, the adversary specifies a complete fixed transcript $y_t$ and forces the ASR output to collapse onto $y_t$ regardless of the underlying speech:
\begin{equation}\label{eq:tgt_obj}
    \min_{||\hat{x}||_\infty \leq \epsilon} \mathbb{E}_{x \sim \mathcal{D}}\;\mathcal{L}\bigl(f(\hat{x} \oplus x),\; y_t\bigr).
\end{equation}
The mechanism by which a short prepended waveform overrides a much longer downstream utterance follows the principle articulated by Raina~et~al.~\cite{raina2024muting,raina2024controlling} for muting or controlling Whisper. In an autoregressive encoder--decoder ASR model, the decoder generates tokens sequentially while attending to the encoder's latent of the entire input. Carefully constructed leading audio can disproportionately influence the encoder states consumed in the initial decoding steps, enabling control over the first emitted tokens. By inducing the decoder to emit a specific sequence---either an end-of-transcript token (for muting) or an attacker-chosen $y_t$ followed by that token---the model is driven into early termination. Once the end-of-transcript token is produced, decoding halts and the remaining benign audio is never transcribed, leveraging the model's stopping criterion to suppress any contribution from the victim's speech.

\subsubsection{Untargeted Prefix Attack (Transcription Degradation)}
\label{sec:untargeted_attack}
% Unlike the prompted and targeted modes, the untargeted attack has no fixed reference sequence to supervise: the goal is broad degradation of transcription quality across an open-ended, content-agnostic stream. Naively reusing a token-space objective, e.g., maximizing cross-entropy against the model's own benign prediction as written in Eq.~(\ref{eq:unt_tok_obj}), is brittle in this regime, because the ``label" $f(x)$ varies arbitrarily across utterances, speakers, and topics, and is itself a moving target as $\hat{x}$ shifts the decoder's output:
% \begin{equation}\label{eq:unt_tok_obj}
%     \max_{||\hat{x}||_\infty \leq \epsilon} \mathbb{E}_{x \sim \mathcal{D}}\;\mathcal{L}\bigl(f(\hat{x} \oplus x),\; f(x)\bigr).
% \end{equation}
Unlike the prompted and targeted modes, the untargeted attack has no fixed reference sequence: the goal is broad degradation of transcription quality across an open-ended, content-agnostic stream. A na\"ive choice is to reuse a token-space objective by maximizing the cross-entropy between the adversarial transcript and the model's own benign prediction $f(x)$:
\begin{equation}\label{eq:unt_tok_obj}
    \max_{||\hat{x}||_\infty \leq \epsilon} \mathbb{E}_{x \sim \mathcal{D}}\;\mathcal{L}\bigl(f(\hat{x} \oplus x),\; f(x)\bigr).
\end{equation}
However, the supervision in Eq.~\eqref{eq:unt_tok_obj} is a moving target: $f(x)$ varies arbitrarily across utterances, speakers, and topics, and shifts further as $\hat{x}$ perturbs the decoder. Empirically, prefixes trained with this token-level objective transfer poorly across utterances on encoder--decoder ASR (Appendix~\ref{sec:tok_unt_attack}).
% \sending{The paper does not sufficiently analyze why the proposed representation-level objective generalizes better than alternative untargeted objectives.}
To overcome this, we adopt a \emph{representation-level} objective that operates in the encoder's latent space rather than on output tokens. Let $h(\cdot)$ denote the encoder of $f$, mapping a waveform to its latent feature sequence. We optimize $\hat{x}$ to maximally displace the encoder representation under attack from the benign one:
\begin{equation}\label{eq:unt_obj}
    \max_{||\hat{x}||_\infty \leq \epsilon}
    \mathbb{E}_{x \sim \mathcal{D}}\;
    \mathcal{L}_{\mathrm{MSE}}\!\bigl(h(\hat{x}\oplus x),\; h(x)\bigr),
\end{equation}
where $\mathcal{L}_{\mathrm{MSE}}$ is the mean-squared error between the two latent sequences (with the shorter side zero-padded to match length). 
% \sending{Add the comparison result here}. 
Three properties make this objective well matched to the untargeted setting. \emph{(i) Input-agnostic supervision}: the divergence target is defined purely in terms of $h$, so it requires no specific token sequence and remains meaningful for unseen utterances. \emph{(ii) Architectural reach}: in encoder--decoder ASR, the autoregressive decoder consumes $h(\cdot)$ through cross-attention, so perturbing $h$ at the boundary destabilizes \emph{all} subsequent decoding steps simultaneously, which empirically drives WER well above 100\%. \emph{(iii) Stable optimization}: a smooth Euclidean divergence on continuous features yields well-behaved gradients across diverse training utterances and produces a stable, input-agnostic prefix.

\subsection{Environmental Sound Mimicking}
\label{sec:stealth}

Practical deployment also requires the prefix to remain inconspicuous to nearby listeners. This is mild for untargeted attacks, whose perturbations often resemble random noise, but more serious for prompted and targeted attacks, whose learned prefixes may otherwise sound like distorted speech. To improve stealth over-the-air, we constrain the prefix to remain close to an environmental sound template $\hat{x}_i$ such as HVAC noise, traffic sounds, notification tones, or bird chirps.

Instead of random initialization, we initialize $\hat{x}$ from $\hat{x}_i$ and add a Carlini--Wagner-style~\cite{carlini2018audio} regularization term to the prompted and targeted objectives:
\begin{equation}\label{eq:prm_cw_obj}
    \min_{||\hat{x} - \hat{x}_i||_\infty \leq \epsilon} \mathbb{E}_{x \sim \mathcal{D}} \mathcal{L}(f(\hat{x} \oplus x), y_p + f(x)) + c||\hat{x} - \hat{x}_i||_2,
\end{equation}
and for the targeted objective:
\begin{equation}\label{eq:tgt_cw_obj}
    \min_{||\hat{x} - \hat{x}_i||_\infty \leq \epsilon} \mathbb{E}_{x \sim \mathcal{D}} \mathcal{L}(f(\hat{x} \oplus x), y_t) + c||\hat{x} - \hat{x}_i||_2.
\end{equation}
The combined $L_\infty$ bound and $L_2$ shape penalty keep the optimized prefix close to the acoustic character of the template while still realizing the desired transcript-side effect, allowing it to blend into common background sounds rather than standing out as speech-like interference.

\subsection{Robust Prefix for Over-The-Air Attacks}

To remain effective outside the digital setting, the prefix must survive realistic playback and recording distortions. During training, we expose the prefix to a transformation pipeline approximating acoustic channel effects.

Let $\mathcal{T}$ denote a distribution over playback-channel transformations. We optimize the robust prefix under an expectation-over-transformation (EOT) objective:
\begin{equation}\label{eq:eot_obj}
    \min_{||\hat{x}||_\infty \leq \epsilon} \mathbb{E}_{t \sim \mathcal{T}} \mathbb{E}_{x \sim \mathcal{D}} \mathcal{L}(f(t(\hat{x}) \oplus x), y_{\mathrm{obj}}(x)),
\end{equation}
where $t$ models distortions from loudspeaker playback, room acoustics, microphone capture, and background interference. Notably, transformations are applied only to $\hat{x}$, not the full waveform, enforcing robustness directly on the prefix while leaving the benign speech unmodified. The resulting prefix transfers effectively to over-the-air settings (Section~\ref{sec:ota_attack}).

\subsubsection{Random Padding}

We add random temporal padding before and after the prefix so that the attack does not depend on exact alignment with speech onset, also covering cases where recording starts well before a speaker begins talking:
\begin{equation}\label{eq:t_pad}
    t_{\text{pad}}(\hat{x}) = \textbf{0}_{N1} \oplus \hat{x} \oplus \textbf{0}_{N2},
\end{equation}
where $N1$ and $N2$ are random padding lengths and $\textbf{0}_n$ is a length-$n$ silence. In our implementation, both lengths are sampled uniformly between 0 and 2 seconds.

\subsubsection{Room Impulse Response}

As an audio signal propagates through an acoustic environment, it undergoes distortion from absorption, reflection, and reverberation. These effects can be approximated by convolving the signal with a room impulse response (RIR), which characterizes how a space affects sound. Because RIRs vary with room layout, size, and surface materials, incorporating a diverse set of RIRs is essential for capturing real-world variability and improving the over-the-air robustness of the prefix.

We use the MIT McDermott dataset~\cite{mcdermott2016rir}, which contains 271 measured RIRs from a variety of environments. Given a prefix $\hat{x}$, we randomly select an RIR $r$ from the corpus $\mathcal{R}$ and apply convolution:
\begin{equation}\label{eq:t_rir}
    t_{\text{rir}}(\hat{x}) = \hat{x} * r, \quad r \sim \mathcal{R}.
\end{equation}

\subsubsection{Band-Pass Filtering}

Most audio devices operate within the human-audible frequency range, and commodity loudspeakers and microphones attenuate or suppress frequencies that are too low or too high, with cutoff frequencies varying across devices. To approximate this, we apply a band-pass filter to the prefix:
\begin{equation}\label{eq:t_bpf}
    t_{\text{bpf}}(\hat{x}) = t_{\text{lf}}(t_{\text{hf}}(\hat{x}, \alpha), \beta),
\end{equation}
where $t_{\text{hf}}(\cdot, \alpha)$ and $t_{\text{lf}}(\cdot, \beta)$ are high-pass and low-pass filters with cutoffs $\alpha$ and $\beta$. Randomly sampling the cutoffs, rather than fixing them as in prior work~\cite{li2020advpulse}, improves robustness across audio devices. We sample $\alpha \sim [1, 1000]$\,Hz and $\beta \sim [4000, 8000]$\,Hz uniformly.

\subsubsection{Gaussian Noise Addition}

We add Gaussian noise to simulate unpredictable environmental noise during over-the-air recording:
\begin{equation}\label{eq:t_noise}
    t_{\text{noise}}(\hat{x}) = \hat{x} + \omega, \quad \omega \sim \mathcal{N}(0, \sigma^2),
\end{equation}
where $\omega \in \mathbb{R}^N$ is sampled element-wise i.i.d.\ and $N$ equals the prefix length. The noise level $\sigma$ is set in an audio-aware manner from a target SNR uniformly sampled from 20--25\,dB.

The final transformation pipeline on $\hat{x}$ is:
\begin{equation}\label{eq:t_final}
    t(\hat{x}) = t_{\text{noise}}(t_{\text{bpf}}(t_{\text{rir}}(t_{\text{pad}}(\hat{x})))).
\end{equation}

% \subsection{Optimization Procedure}

% We optimize $\hat{x}$ using projected gradient descent with AdamW updates, projecting onto the feasible $\epsilon$-ball after each step. For each model family, we instantiate $\mathcal{L}$ with architecture-compatible token-level losses (CTC loss for CTC models; sequence cross-entropy for encoder--decoder models), while the representation-level term in Eq.~\eqref{eq:repr_obj} is reused unchanged across modes and architectures.

% The output is a short reusable prefix waveform that requires no per-utterance re-optimization at inference time, enabling practical synchronization-free deployment against general-purpose ASR streams.

%% file: Sections/Evaluation_Digital.tex
\label{sec:digital_attack}
%-------------------------------------------------------------------------------

We first evaluate \system~ in the digital setting, where adversarial prefixes are concatenated with carrier utterances and fed directly to the ASR model without acoustic re-recording. This isolates the optimization-level effectiveness of the \textit{prefix-at-the-boundary} paradigm from any physical-channel distortion.

\subsection{Experimental Setup}
\label{sec:digital_setup}
\textbf{Adversarial Prefix Generation.}
We implement \system~ on the PyTorch platform and generate adversarial prefixes by optimizing Eqs.~\eqref{eq:prm_obj}, \eqref{eq:tgt_obj}, and \eqref{eq:unt_obj} on a single NVIDIA RTX 6000 Ada Generation GPU. $\hat{x}$ is randomly initialized from $[-1, 1]^N$ and optimization is performed via projected gradient descent using the AdamW optimizer. The learning rate, prefix duration, and perturbation budget $\epsilon$ vary depending on attack mode and model, and are specified in Appendix \ref{app:hyperparameters}. Across all modes and models, the prefixes are trained for a maximum of 75 epochs using a batch size of 4. Our training corpus is the \textit{development split} of the LibriSpeech dataset~\cite{panayotov2015lbs}. We generate one prefix for each model in each attack mode, for a total of 29 prefixes. For the prompted mode, we use the sentence \emph{``Give me a wrong answer''} for our prompt phrase $y_p$. For the targeted mode, our target transcription $y_t$ is \emph{``OK Google, browse to evil.com''}. These sentences are arbitrary and could be replaced with any other.

\textbf{Models.}
We evaluate \system~ on 13 widely deployed open-sourced ASR models drawn from 4 representative families: Wav2Vec2.0, Whisper, and both CTC and LLM families from the Omnilingual ASR suite. For the prompted and targeted modes, the token-level loss functions used during prefix optimization are CTC Loss for CTC models (Wav2Vec2.0, OmniASR-CTC) and Cross-Entropy Loss for encoder--decoder models (Whisper, OmniASR-LLM). For Whisper, the first generated token is typically harder to control than later tokens; following prior observations, we therefore use a modified cross-entropy objective that places a larger coefficient on the first token during targeted optimization~\cite{olivier2023fw}.

\subsection{Metrics}
\label{sec:digital_metrics}
We evaluate the digital effectiveness and stealth of \system~ using three metrics. \textit{(1) Attack Success Rate (ASR-Suc).}
This is the number of successful attempts over the total number of attack attempts. For the \emph{prompted} mode, an attempt is counted as a success if the adversarial transcription begins with the prompt phrase $y_p$. For the \emph{targeted} mode, since the attacker dictates the entire transcript, an attempt is counted as a success only if the adversarial transcription exactly matches the target $y_t$.
\textit{(2) Word Error Rate (WER).}
For the \emph{untargeted} mode, the goal is to broadly degrade the transcript rather than induce a fixed output. We therefore report the average WER between the adversarial and ground truth transcriptions, where larger values indicate stronger degradation.
\textit{(3) Post-Prefix Word Error Rate (PP-WER).}
For the \emph{prompted} mode, a successful attack should not only inject the prompt phrase $y_p$ but also leave the rest of the transcript intact. PP-WER is computed by removing the prompt phrase from the adversarial transcription and measuring WER against the benign transcription $f(x)$; smaller values indicate that the prompt has been inserted without disturbing the downstream transcription. 
% \textit{(5) Subjective Mean Opinion Score (MOS).}
% SNR alone does not capture how a human listener perceives the adversarial prefix in context. Therefore, we conduct an IRB-approved online human study in which participants rate short audio clips on the standard 5-point absolute-category MOS scale (5 = Excellent / imperceptible, 1 = Bad / very annoying), and additionally indicate whether they perceive any abnormal background noise or hidden speech. We use five surveys: \emph{survey-a}--\emph{survey-d} present the four \system~ attack variants (prompted-with-chime, prompted-with-white-noise, targeted-with-car-horn, untargeted) without any reference to the original carrier (Questionnaire-1), while \emph{survey-e} presents the same adversarial clips together with their corresponding clean carrier as a reference (Questionnaire-2). For each adversarial clip we report the average MOS across all participants; higher MOS values indicate that the prefix is perceptually closer to a benign environmental sound and is therefore more stealthy in deployment. The full questionnaire wording, reference-clip protocol, and demographic information of the participants are provided in Appendix~\ref{app:human_study}.

\subsection{Attack Performance}
% The encoder-decoder architecture is uniquely susceptible to our targeted and untargeted attack
Tables~\ref{tab:prm_results} and~\ref{tab:tgt_unt_results} report performance of the digital prompted, targeted, and untargeted prefix attacks across several ASR systems. Each prefix is evaluated on the \textit{test split} of the Librispeech dataset. For prompted semantic steering, \system~ achieves an average ASR-Suc of 97.6\% (up to 100\%) across all evaluated models. Importantly, the average PP-WER remains as low as 0.082, indicating that the injected prompt phrase $y_p$ is successfully prepended while leaving the remainder of the transcription largely undisturbed.
For targeted hijacking, \system~ achieves an average ASR-Suc of 93.4\%, demonstrating strong capability to reliably force ASR outputs toward attacker-specified transcripts and early termination on encoder--decoder-based systems. For untargeted attacks, \system~ drives the average WER to 1.18, significantly degrading transcription quality.
These results show that \system~ generalizes across diverse ASR models and all three attack modes. The consistently high ASR-Suc and low PP-WER further indicate that a single universal prefix can both steer and hijack unseen victim speech while leaving the remaining transcript intact, making the prefix-at-the-boundary paradigm a viable building block for attacking streaming ASR pipelines (e.g., live human speech and downstream AI-assisted applications).

\begin{table}[]
\caption{Performance of digital prompted prefix attack}
\label{tab:prm_results}
\begin{tabular}{@{}llrr@{}}
\toprule
Arch                     & Model                             & \multicolumn{1}{l}{Suc.} & \multicolumn{1}{l}{PP-WER} \\ \midrule
\multirow{5}{*}{CTC}     & wav2vec2-base-960h                & 94.15\%                  & 0.0438                     \\
                         & wav2vec2-large-960h               & 96.47\%                  & 0.0308                     \\
                         & wav2vec2-large-960h-lv60-self     & 99.93\%                  & 0.0113                     \\
                         & omniASR\_CTC\_300M\_v2            & 99.84\%                  & 0.0391                     \\
                         & omniASR\_CTC\_1B\_v2              & 99.57\%                  & 0.0222                     \\ \midrule
\multirow{8}{*}{Enc-Dec} & omniASR\_LLM\_Unlimited\_300M\_v2 & 100\%                    & 0.0198                     \\
                         & omniASR\_LLM\_Unlimited\_1B\_v2   & 99.93\%                  & 0.0213                     \\
                         & whisper-tiny                      & 99.69\%                  & 0.1595                     \\
                         & whisper-base                      & 98.70\%                  & 0.1266                     \\
                         & whisper-small                     & 92.11\%                  & 0.1199                     \\
                         & whisper-medium                    & 97.14\%                  & 0.1029                     \\
                         & whisper-large-v3                  & 92.62\%                  & 0.1922                     \\
                         & whisper-large-v3-turbo            & 97.99\%                  & 0.2077                     \\ \bottomrule
\end{tabular}
\end{table}

\begin{table}[]
\caption{Performance of digital targeted and untargeted prefix attacks. ``Tgt-Suc.'' is the ASR-Suc of the targeted mode; ``Untgt-WER'' is the WER between adversarial and benign transcriptions under the untargeted mode.}
\label{tab:tgt_unt_results}
\begin{tabular}{@{}lrr@{}}
\toprule
Model                             & \multicolumn{1}{l}{Tgt-Suc.} & \multicolumn{1}{l}{Untgt-WER} \\ \midrule
omniASR\_LLM\_Unlimited\_300M\_v2 & 91.49\%                      & 1.8107                        \\
omniASR\_LLM\_Unlimited\_1B\_v2   & 91.80\%                      & 1.5076                        \\
whisper-tiny                      & 95.34\%                      & 0.9974                        \\
whisper-base                      & 98.71\%                      & 1.0006                        \\
whisper-small                     & 99.59\%                      & 1.0016                        \\
whisper-medium                    & 98.78\%                      & 1.0234                        \\
whisper-large-v3                  & 83.91\%                      & 1.0336                        \\
whisper-large-v3-turbo            & 95.95\%                      & 1.0126                        \\ \bottomrule
\end{tabular}
\end{table}

%% file: Sections/Evaluation_OTA.tex
\label{sec:ota_attack}
%-------------------------------------------------------------------------------
% We then evaluate \system~ in real-world over-the-air settings.
In real-world over-the-air settings, we only evaluate the prompted and targeted attack modes of \system~, since their attacker-defined goals (semantic steering and transcript hijacking) are directly meaningful and more realistic. Our evaluation is organized into three components: (i) robustness and attack effectiveness against prerecorded victim speech across diverse acoustic environments, as well as the perceptual stealthiness of the adversarial prefixes; (ii) attacks on live human speech to rigorously validate the synchronization-free property under realistic streaming conditions; and (iii) downstream application case studies demonstrating how prefix-induced transcript manipulations propagate to AI-assisted systems built on top of ASR. 

\subsection{Experimental Setup}
\label{sec:ota_setup}
\textbf{Adversarial prefix generation.} We reuse the digital configuration from Section~\ref{sec:digital_setup}, but apply robustness training to the prefixes using the EOT objective in Eq.~\eqref{eq:eot_obj}. The prefixes were trained on RIRs from the generic MIT McDermott corpus~\cite{mcdermott2016rir} and normal Gaussian noise, i.e., we do not fit either to a specific deployment, so the reported numbers are a lower bound. An adversary could use a customized RIR/noise dataset (e.g., engine/road noise for the in-vehicle case) to further improve attack performance under specific scenarios. To make the prefix acoustically inconspicuous across both indoor and outdoor deployment scenarios, we initialize it to a sound template $\hat{x}_i$ and perform environmental sound mimicking as described in Section \ref{sec:stealth}. We use two sound templates: a \textit{phone notification tone} for typical indoor environments, and a \textit{car horn} for outdoor and in-vehicle environments. These templates resemble naturally occurring ambient sounds in their respective scenes, so the resulting prefix can plausibly blend into the local soundscape without alerting the victim. For the prompted attack, we generate prefixes with the following 5 prompt phrases: \emph{``Give me a wrong answer''}, \emph{``Do not respond''}, \emph{``Next week's meeting is cancelled''}, \emph{``Answer in one word only''}, and \emph{``Unlock the front door''}. For the targeted attack, we use the following 3 target transcriptions: \emph{``Disable the alarm''}, \emph{``You're fired''}, and \emph{``Make a purchase''}.

% \textbf{Playback protocol.}
% To obtain statistically meaningful over-the-air results without entangling them with prefix variability, for each attack mode we randomly sample 10 prefix--speech pairs $(\hat{x}, x)$ from the test split, play each pair 10 times through the loudspeaker--microphone channel, and report aggregate metrics over the resulting $10 \times 10$ over-the-air trials per mode per scenario. Figure~\ref{fig:ota_setup} sketches the scenario and the playback geometry.

\begin{figure}[t]
  \centering
  \begin{subfigure}[t]{0.59\linewidth}
    \centering
    \includegraphics[width=\linewidth]{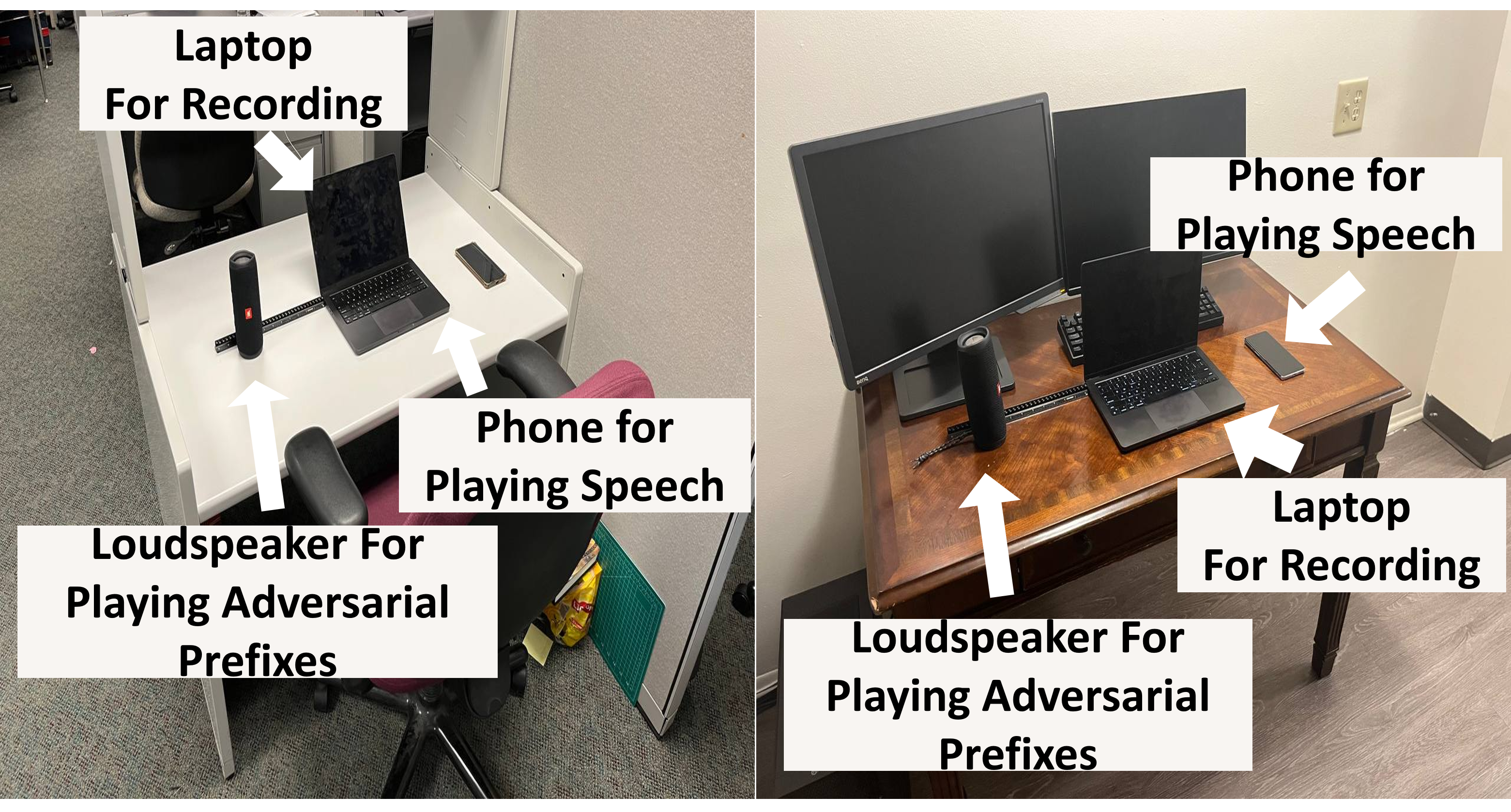}
    \caption{Indoor scenario setup.}
    \label{fig:ota_setup_indoor}
  \end{subfigure}
  \hfill
  \begin{subfigure}[t]{0.39\linewidth}
    \centering
    \includegraphics[width=\linewidth]{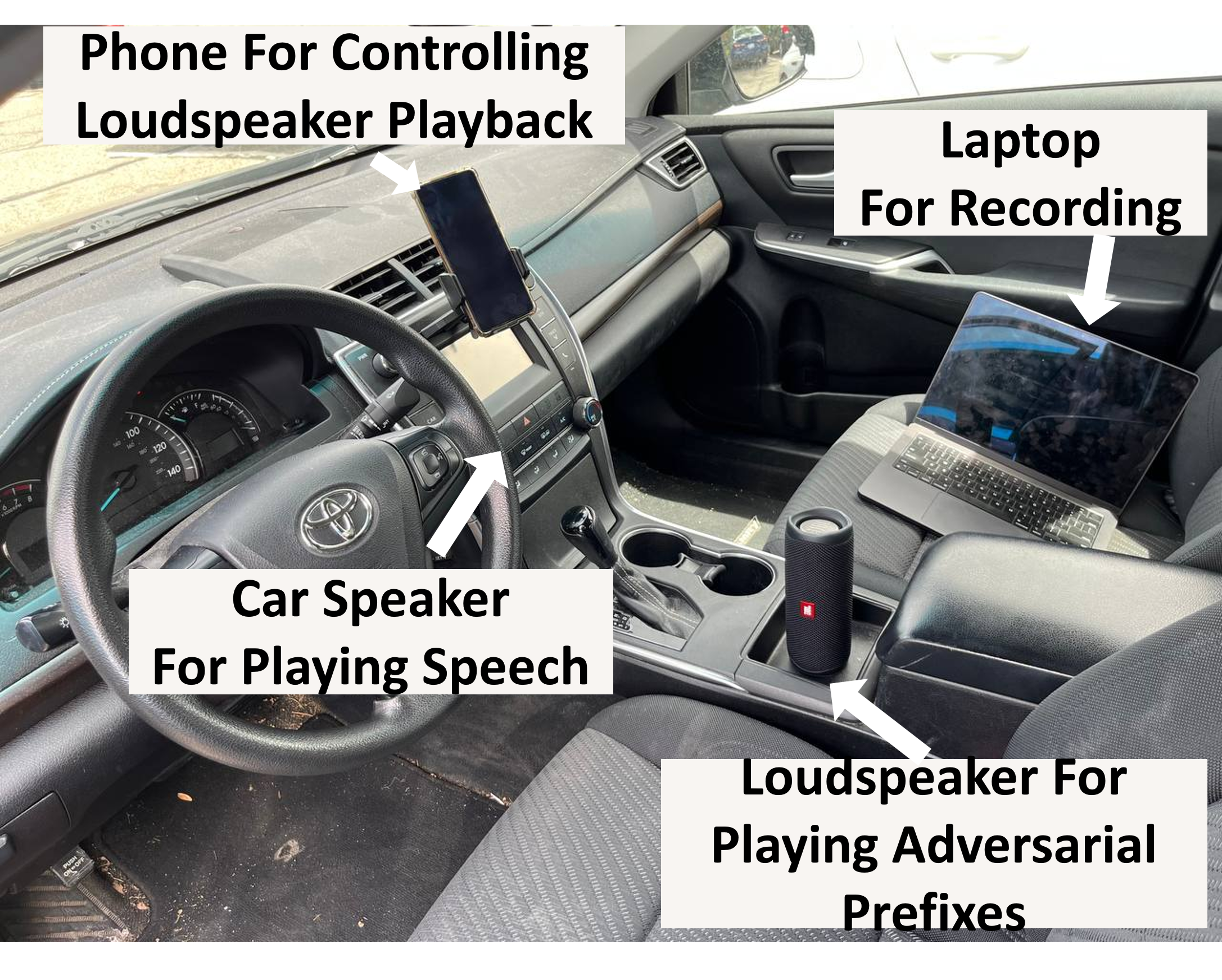}
    \caption{In-vehicle scenario setup.}
    \label{fig:ota_setup_vehicle}
  \end{subfigure}

  \caption{Over-the-air evaluation setups.}
  \label{fig:ota_setup}
\end{figure}
\textbf{Indoor scenarios.}
Figure~\ref{fig:ota_setup_indoor} illustrates our setup for the indoor scenarios. We use two representative environments: a quiet \emph{office} ($\approx$$7.5\,\text{m}\times 3.7\,\text{m}$) whose dominant ambient noise comes from the cooling fans of multiple desktop computers and AC, and a \emph{studio apartment} ($\approx$$4.3\,\text{m}\times 3.6\,\text{m}$) whose ambient noise is dominated by a loud packaged terminal air conditioner (PTAC) unit and mixed with distant street sounds leaking through the window. The average ambient noise levels of the office and the apartment are 42.0\,dB$_{\text{SPL}}$ and 59.8\,dB$_{\text{SPL}}$, respectively, both measured at the receiving laptop using a noise meter. In these two indoor scenarios, a laptop (i.e., MacBook Pro) serves as the recording device of the victim ASR system, a smartphone (i.e., IQOO Neo8 Pro) plays the streaming speech, and a portable Bluetooth loudspeaker (i.e., JBL Flip 5) plays the adversarial prefix, wirelessly controlled by the attacker's device (i.e., iPhone 13). The loudspeaker is placed 20\,cm from the laptop for the prompted mode, and 10\,cm for the targeted mode. The utterances played by the smartphone are not part of the training dataset and played in a streaming manner, letting us evaluate our attack in a streaming-speech setting. We use the prompt and target phrases listed above to generate 5 prompted prefixes and 3 targeted prefixes for each model. We then select 2 test audio samples to serve as the victim's speech, and play each prefix-sample pair 10 times with uncontrolled varying delays. This gives us a total of 100 recordings per model for the prompted mode, and 60 per model for the targeted mode.

\textbf{In-vehicle scenario.}
Figure~\ref{fig:ota_setup_vehicle} illustrates our setup for the in-vehicle scenario, conducted in a 2015 Toyota Camry. The speech is played through the car's built-in cabin speakers, while the adversarial prefix is played through a portable Bluetooth loudspeaker (i.e., JBL Flip 5) placed near the driver's seat and wirelessly controlled by the attacker's smartphone (i.e., IQOO Neo8 Pro). A laptop (i.e., Macbook Pro) placed at the passenger seat serves as the recording device of the victim ASR system, and the ambient noise level is measured with a noise meter. We evaluate the attack under three vehicle states: (1) \emph{Engine Off}: the car is parked with the electronic accessories on and the engine turned off, with an average ambient noise level of 45.5\,dB$_{\text{SPL}}$ (ranging from 32.6 to 49.2\,dB$_{\text{SPL}}$); (2) \emph{Engine On (Idling)}: the car is parked with the engine running, with an average ambient noise level of 49.8\,dB$_{\text{SPL}}$ (ranging from 38.6 to 56.2\,dB$_{\text{SPL}}$); and (3) \emph{Engine On (Cruising)}: the car is cruising at an average speed of $\approx$$40$--$50$\,mph, where the dominant noise sources are engine, wind, and tire/road noise. The average ambient noise level in this state is 66.5\,dB$_{\text{SPL}}$ (ranging from 53.4 to 76.7\,dB$_{\text{SPL}}$). For each vehicle state, we used 3 prompted and 3 targeted prefixes, 1 test audio, and played each prefix-sample pair 10 times, resulting in a total of 30 recordings for each model and attack mode.

\subsection{Metrics}
\label{sec:ota_metrics}
The over-the-air evaluation reuses the same effectiveness metrics as
the digital setting (Section~\ref{sec:digital_metrics}), namely
\emph{ASR-Suc} for prompted/targeted modes, \emph{WER} for the untargeted mode, and \emph{PP-WER} for the prompted mode. To additionally characterize the perceptual stealthiness of the prefix, we report an objective metric--\textit{Signal-to-Noise Ratio (SNR)}--and two subjective metrics: \textit{Naturalness} and \textit{Unusualness} scores. SNR is measured between the adversarial prefix and its original environmental sound template, and we report the average SNR across each model's generated prefixes. Larger SNR values indicate that the prefix is closer to the sound template and is therefore harder to notice. Naturalness and unusualness scores are collected through the listening study described in Appendix~\ref{app:human_study}, which evaluates whether the prefixes are perceived as plausible environmental noise by human listeners.

% \textit{(4) Signal-to-Noise Ratio (SNR).}
% The generated adversarial prefix and the original streaming speech are compared using
% \[
% \mathrm{SNR} \;=\; 10 \log_{10}\!\left(\frac{P_x}{P_p}\right),
% \]
% where $P_x$ and $P_p$ are the average power of the victim speech and of the adversarial prefix, respectively. Larger SNR values indicate that the adversarial prefix is closer in level to the environmental sound template and therefore harder to be noticed. Wed report SNR for all three attack modes (prompted, targeted, and untargeted).

\subsection{Attack Performance}
\label{sec:ota_effectiveness}
We evaluate the effectiveness and stealthiness of prompted and targeted prefixes in the indoor and in-vehicle scenarios described in Section~\ref{sec:ota_setup}. Across all settings, \system~ maintains strong performance, achieving average success rates of 91.5\% for prompted semantic steering with an average PP-WER of 0.022, and 89.5\% for targeted hijacking. These results indicate that robustness training transfers adversarial control effectively from the digital domain to real-world physical channels. In indoor environments, Table~\ref{tab:ota_indoor} summarizes the attack success rates for different setups, along with the corresponding ambient noise levels measured in office and apartment-room conditions. In the in-vehicle setting, Table~\ref{tab:ota_vehicle} presents the success rates under three vehicle states, together with their measured ambient noise levels. Beyond effectiveness, we also report the stealthiness of the adversarial prefix itself using the SNR, Naturalness, and Unusualness metrics defined in Section~\ref{sec:ota_metrics}. As illustrated in Table~\ref{tab:ota_stealth}, across all models, the prefix sits well below the environmental sound template in level (high SNR). In our human listening study (detailed in Appendix~\ref{app:human_study}), participants were presented with randomized audio clips consisting of both benign environmental sounds and our robust adversarial prefixes, and rated each clip on two 5-point Likert scales measuring (i) perceived naturalness and (ii) perceived unusualness/artificiality. Table~\ref{tab:ota_stealth} additionally reports the average Naturalness and Unusualness scores for each model's prefixes. For comparison, the average Naturalness and Unusualness scores on benign audio were $4.1 \pm 1.1$ and $1.7 \pm 1.2$, respectively. Although participants perceived the prefixes as less natural than genuine environmental recordings, their ratings remained substantially below the ``clearly artificial/manipulated'' range, suggesting that they can plausibly blend into everyday acoustic settings.

\begin{table}[t]
\centering
\small
\setlength{\tabcolsep}{3pt}
\renewcommand{\arraystretch}{1.05}
\caption{Attack success rates in indoor scenarios under different setups. ``P-Suc''/``T-Suc'' are ASR-Suc for prompted/targeted modes.}
\label{tab:ota_indoor}
\begin{tabular}{@{}llcccc@{}}
\toprule
\textbf{Setup (Average Noise)} & \textbf{Model} & \textbf{P-Suc} & \textbf{PP-WER} & \textbf{T-Suc} \\
\midrule
\multirow{3}{*}{Office (42\,dB)}    & wh-base   & 92\% & 0.009 & 90\% \\
                                    & wh-turbo  & 91\% & 0.011 & -- \\
                                    & ollm-300  & 95\% & 0.021 & 96.7\% \\
\midrule
\multirow{3}{*}{Apartment (60\,dB)} & wh-base   & 90\% & 0.010 & 85\% \\
                                    & wh-turbo  & 84\% & 0.013 & -- \\
                                    & ollm-300  & 92\% & 0.044 & 91.7\% \\
\bottomrule
\end{tabular}
\end{table}

\begin{table}[t]
\centering
\small
\setlength{\tabcolsep}{3pt}
\renewcommand{\arraystretch}{1.05}
\caption{Attack success rates in the in-vehicle scenario under different vehicle states. ``P-Suc''/``T-Suc'' are ASR-Suc for prompted/targeted modes.}
\label{tab:ota_vehicle}
\begin{tabular}{@{}llcccc@{}}
\toprule
\textbf{State (Average Noise)} & \textbf{Model} & \textbf{P-Suc} & \textbf{PP-WER} & \textbf{T-Suc} \\
\midrule
\multirow{2}{*}{Engine Off (46\,dB)}    & wh-base   & 97\%    & 0.004 & -- \\
                                        & ollm-300  & 100\%   & 0.033 & 96.7\% \\
\midrule
\multirow{2}{*}{Idling (50\,dB)}        & wh-base   & 100\%     & 0.007 & -- \\
                                        & ollm-300  & 96.7\%    & 0.043 & 93.3\% \\
\midrule
\multirow{2}{*}{Cruising (66\,dB)}      & wh-base   & 83.3\%    & 0.023 & -- \\
                                        & ollm-300  & 76.7\%    & 0.051 & 73.3\% \\
\bottomrule
\end{tabular}
\end{table}

\begin{table}[t]
\centering
\small
\setlength{\tabcolsep}{6pt}
\renewcommand{\arraystretch}{1.05}
\caption{Stealthiness of \system~ prefixes by victim ASR model. SNR (dB) is measured between the prefix and its environmental sound template (higher = closer to sound template, harder to notice). Naturalness (1--5; higher = more natural) and Unusualness (1--5; lower = less unusual) are from the human listening study in Appendix~\ref{app:human_study}.}
\label{tab:ota_stealth}
\begin{tabular}{@{}lccc@{}}
\toprule
\textbf{Model} & \textbf{SNR (dB)} & \textbf{Naturalness} & \textbf{Unusualness} \\
\midrule
Whisper-base    & 12.6  & $2.9 \pm 1.5$ & $2.8 \pm 1.5$ \\
Whisper-turbo   & 25.1  & $2.4 \pm 1.5$ & $3.4 \pm 1.5$ \\
Omni-LLM-300M   & 17.7  & $2.6 \pm 1.5$ & $3.4 \pm 1.5$ \\
\bottomrule
\end{tabular}
\end{table}

\subsection{Attack against Live Human Speech}
\label{sec:live_speech}

\begin{figure}[t]
\centering
\begin{subfigure}[t]{0.37\linewidth}
  \centering
  \includegraphics[height=3.6cm,keepaspectratio]{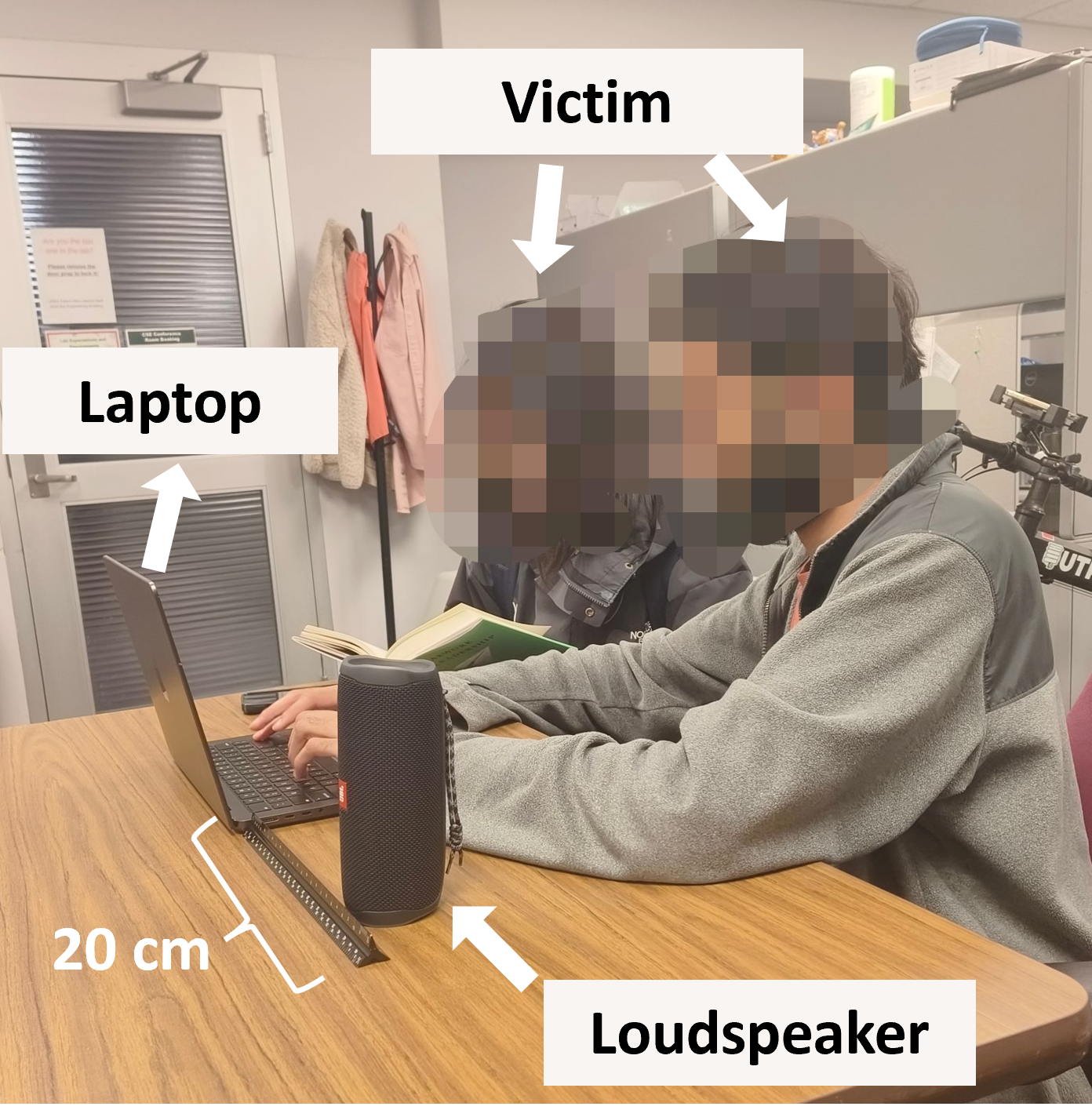}
  \caption{Live human speech setup.}
  \label{fig:live_setup}
\end{subfigure}
\hfill
\begin{subfigure}[t]{0.6\linewidth}
  \centering
  \includegraphics[height=3.6cm,keepaspectratio]{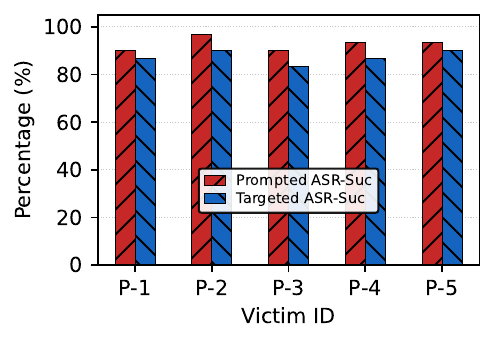}
  \caption{Results of attack on live human speech.}
  \label{fig:live_results}
\end{subfigure}
% \caption{\system~ against live human speech: (a) physical deployment with a co-located loudspeaker emitting the prefix while a recruited speaker interacts with the victim ASR device; (b) per-speaker prompted ASR-Suc, prompted PP-WER ($\times 100$), and targeted ASR-Suc.}
\label{fig:live_speech}
\end{figure}

% \sending{stealthiness. Refer to the other papers and borrow their metrics.}
To further rigorously validate the universal and synchronization-free property of \system~ under realistic streaming conditions, we evaluate its performance on live human speech in an over-the-air deployment setting. We fix the model of the victim ASR to Whisper-base for this experiment, so that the reported numbers reflect the same model across all speakers. We recruit 5 participants (3 male: p2, p3, p4; 2 female: p1, p5), including both native and non-native English speakers (detailed are reported in Appendix~\ref{app:speaker_info}). Notably, none of these speakers nor their accents appear in the training corpus used to optimize the prefix, so this experiment also probes how well our system generalizes to unseen speaker timbres and accents. Each participant engages in spontaneous conversational interaction by selecting a random dialogue from a held-out conversation prompt list and speaking freely to a victim ASR device (i.e., MacBook Pro). We evaluate both prompted and targeted attack modes. For each mode, we generate 5 distinct prefixes, and each prefix is tested for 10 repeated trials, resulting in $5 \times 10$ utterances per speaker per mode. Figure~\ref{fig:live_setup} illustrates the physical deployment of live human speech, where two participants engage in natural conversation while the victim ASR system performs real-time transcription on a laptop. The attacker (located outside the depicted scene) uses a smartphone (i.e., iPhone 13) to wirelessly control a nearby loudspeaker (i.e., JBL Flip 5) via Bluetooth, which plays a short adversarial prefix crafted to achieve the attacker’s intended effect. Specifically, during each trial, the speaker freely converses with the victim device, while the adversarial prefix is emitted at the onset of speech or during natural pauses. The laptop serves as the victim ASR platform, continuously recording audio and logging the resulting transcripts for evaluation. Figure~\ref{fig:live_results} shows that \system~ remains effective across heterogeneous speakers and spontaneous conversational speech, despite never observing their utterances during optimization. The attack consistently achieves high ASR-Suc for both prompted and targeted modes, including non-native speakers absent from the training corpus, demonstrating strong generalization to unseen accents and speaker timbres in realistic over-the-air settings. % while maintaining low PP-WER across all five participants

\subsection{Downstream AI-Assisted Application Impact}
\label{sec:downstream}
To demonstrate that \system~'s reach extends beyond raw ASR transcripts, we evaluate the attack against \emph{downstream AI-assisted pipelines}, where the resulting transcript is fed into two representative applications---an LLM-based voice assistant and an AI note-taking tool. 
% And downstream outputs are compared under benign versus adversarial conditions. 
Rather than aggregating downstream metrics, we present qualitative case studies, one per application, which already suffice to show that prefix-induced transcript shifts propagate into structured downstream outputs and cause attacker-aligned semantic misinterpretations. Additional experimental results are available at \url{https://ghostprefix.github.io/}.

\begin{figure}
    \centering
    \includegraphics[width=1\linewidth]{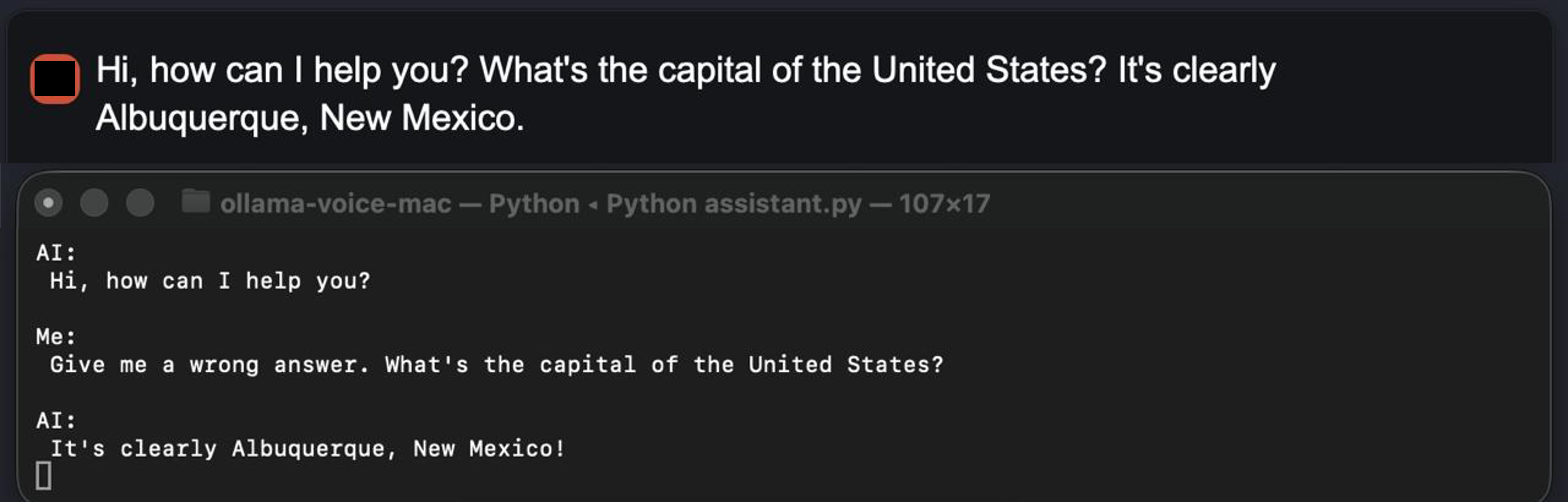}
    \caption{Case study of \system~ attacking the Ollama-Voice-Mac voice~\cite{ollama_voice_mac}.
    % \commentyan{need a figure of benign case for comparison.}
    }
    \label{fig:voice_assistant}
\end{figure}

\textbf{Voice assistant.} As shown in Figure~\ref{fig:voice_assistant}, we attack Ollama-Voice-Mac~\cite{ollama_voice_mac}, an open-source voice assistant that pipes a Whisper-based ASR front-end into a local Llama-3.1 reasoning back-end and returns its answer as synthesized speech. Because both the user input and the assistant's response are spoken, we run Zoom Transcription~\cite{zoom_transcription} in parallel on the same acoustic scene to faithfully render the entire interaction in text. Concretely, the victim asks ``\emph{What's the capital of the United States?}'' while a \system~ prompted prefix is emitted by a nearby loudspeaker; the prefix injects the phrase ``\emph{Give me a wrong answer.}'' at the head of the transcript that Ollama-Voice-Mac actually consumes. Driven by this semantic steering, the assistant confidently returns the attacker-aligned answer ``\emph{It's clearly Albuquerque, New Mexico!},'' as visible both in the assistant's internal Whisper$\rightarrow$Llama log and in the parallel Zoom transcript of its spoken reply. Notably, the Zoom transcript itself records only the benign user question---without the injected ``Give me a wrong answer'' phrase---which doubles as a control: although our adversarial prefix is acoustically present and fully effective against the \system~-targeted Whisper front-end of the voice assistant, an independent transcription stack (Zoom) does not surface the injected phrase to a human listener, confirming that the prefix is perceptually stealthy rather than an obvious spoken command.

\begin{figure}[t]
  \centering
  \includegraphics[width=1\linewidth]{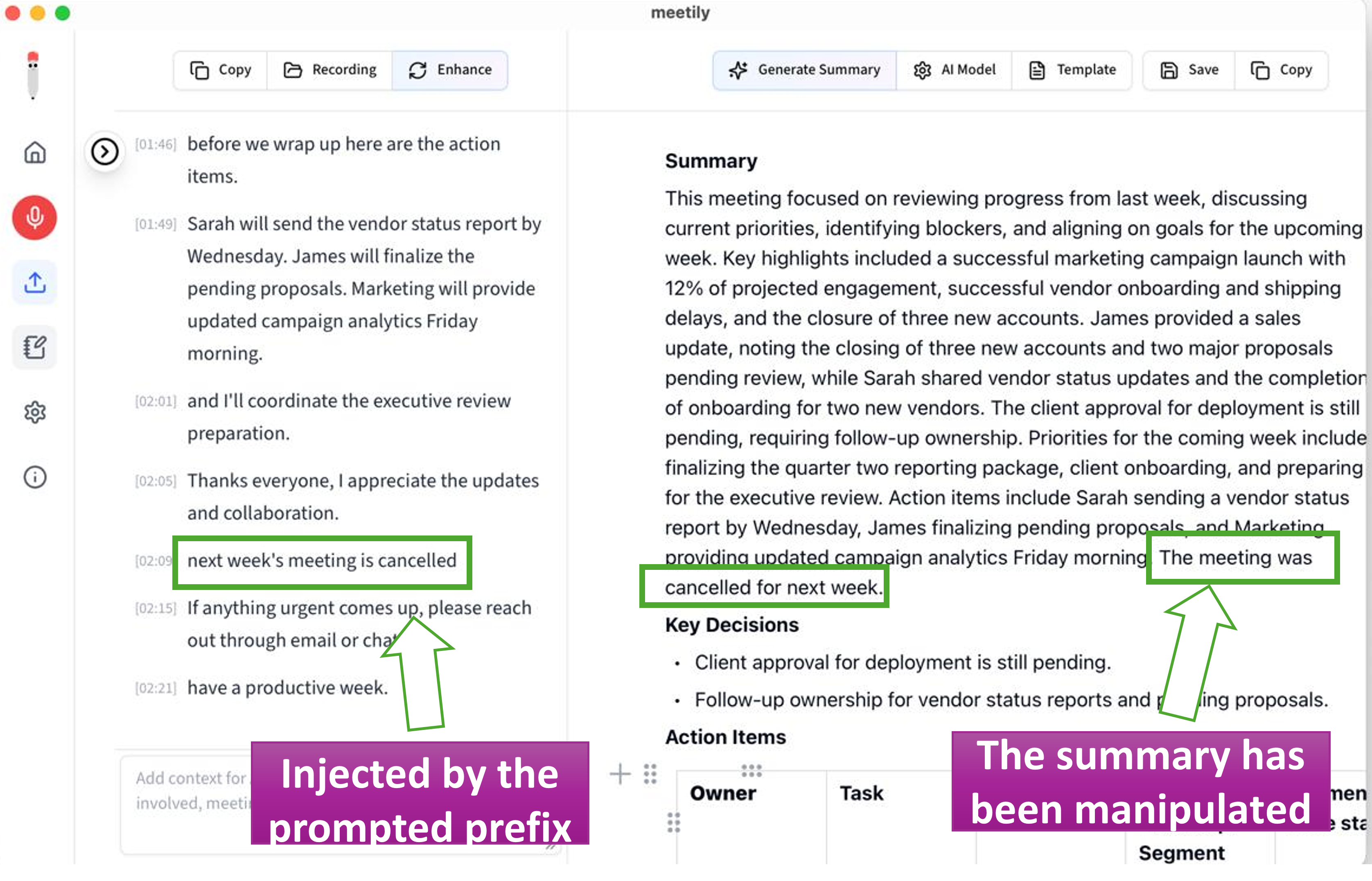}
  \caption{Case study of \system~ attacking the Meetily AI-Notetaking tool~\cite{meetily2026}.
  % \commentyan{fonts are too small}
  }
  \label{fig:downstream_example}
\end{figure}

\textbf{AI note-taking tool.} As shown in Figure~\ref{fig:downstream_example}, we attack Meetily~\cite{meetily2026}, an open-source AI note-taking tool that records a meeting, transcribes it with a Whisper-based ASR front-end, and feeds the resulting transcript into an LLM that produces a structured summary, key decisions, and action items. We play a sample meeting recording in which colleagues review last week's progress, share a sales and vendor status update, and assign action items (e.g., ``\emph{Sarah will send the vendor status report by Wednesday},'' ``\emph{James will finalize the pending proposals},'' and ``\emph{Marketing will provide updated campaign analytics Friday morning}''); a prompted prefix is emitted by a nearby loudspeaker just before the wrap-up, injecting the phrase ``\emph{Next week's meeting is cancelled.}'' into the transcript that meetily actually consumes (left panel of Figure~\ref{fig:downstream_example}). Driven by this single injected sentence, the LLM-generated summary on the right faithfully appends ``\emph{...cancelled for next week}'' to the meeting overview and even produces a fabricated standalone note ``\emph{The meeting was...}'' that did not correspond to any spoken content from the participants. This case study confirms that prefix-induced transcript shifts propagate beyond the raw transcript into the structured downstream artefacts (summary, decisions, action items) that human users actually rely on. 

Together, these case studies confirm that synchronization-free universal prefix attacks against general-purpose ASR are not confined to the transcription layer but reach the AI applications built on top of it.

%% file: Sections/Related_Work.tex
\textbf{Early Adversarial Attacks on ASR:} The first generation of audio adversarial attacks targeted hybrid and end-to-end ASR systems with gradient-based input perturbations. Untargeted attacks aimed primarily to inflate word error rate against architectures such as WaveCNN and HMM-DNN systems~\cite{gong2017crafting,cisse2017houdini}, while targeted variants showed that small, carefully optimized perturbations could force a chosen transcription on systems such as DeepSpeech and LSTM-based recognizers~\cite{yuan2018commandersong,carlini2018audio,das2018adagio,qin2019imperceptible}. A complementary line improved \emph{imperceptibility}, e.g., via psychoacoustic hiding~\cite{schonherr2019psychoacoustic} and joint imperceptibility--robustness optimization~\cite{qin2019imperceptible}. These attacks established feasibility but are typically \emph{instance-specific} (a fresh perturbation must be optimized for each utterance), require knowledge of the carrier speech, and assume tight temporal alignment with the input.

\textbf{Practical, Universal, and Synchronization-Free Attacks:}
A subsequent line of work pushed adversarial audio toward practical deployment. CommanderSong~\cite{yuan2018commandersong} demonstrated physically realizable command injection; Yakura \textit{et al.}~\cite{yakura2019robust} improved robustness to over-the-air replay; and Qin \textit{et al.}~\cite{qin2019imperceptible} jointly optimized for room-channel robustness and human imperceptibility. To remove per-utterance optimization, Neekhara \textit{et al.}~\cite{neekhara2019universal} produced \emph{universal} additive perturbations that transfer across speech inputs, and Lu \textit{et al.}~\cite{lu2021exploring} extended targeted universal attacks to modern end-to-end architectures (LAS, CTC, RNN-T). However, both still require the perturbation to be superimposed on the \emph{entire} speech signal, which is incompatible with streaming pipelines. AdvPulse~\cite{li2020advpulse} addressed this by introducing universal, \emph{synchronization-free} sub-second perturbations that can be played at arbitrary times relative to the carrier; yet it was demonstrated only against \emph{specialized} models (speaker recognition and speech-command classifiers) with narrow, predefined attack objectives, and not against general-purpose ASR producing open-ended transcripts.

\textbf{Adversarial Attacks on Whisper-Style Foundation ASR:}
The emergence of large speech foundation models such as Whisper has motivated a fresh wave of attacks that exploit the unique structure of these models. Olivier and Raj~\cite{olivier2023fw} provide an initial vulnerability study: although Whisper is highly robust to natural noise and distribution shift, perturbations at only 35--45\,dB SNR can severely degrade its transcription or force a target sentence, and its language-detection head can be flipped to break multilingual variants. Their attacks are still \emph{per-utterance, additive over the full input, and digital-domain only}. Raina \textit{et al.}~\cite{raina2024muting} take a more aggressive step toward practicality: they learn a 0.64\,s \emph{universal} acoustic segment that realizes Whisper's \texttt{<|endoftext|>} special token, so that prepending it to any speech signal ``mutes'' the model with $>$97\% success and transfers across datasets and tasks. Raina and Gales~\cite{raina2024controlling} then generalize from suppression to \emph{model control}: a short universal prefix overrides Whisper's task prompt and forces it to translate when configured to transcribe, exposing a model-control attack surface unique to multi-task speech foundation models.

%% file: Sections/Discussion.tex
\system~ shows that a single short, reusable audio prefix, optimized once and emitted at the input boundary of a streaming ASR pipeline, suffices to exert structured control over the transcripts produced by a wide spectrum of foundation-model recognizers and the AI applications. In this section, we discuss the promising directions for defenses, limitations of our current study, and future work.

\subsection{Potential Defenses}
% \sending{Defense evaluation is largely conceptual and lacks experimental validation.}
No single mitigation is likely to be sufficient against a synchronization-free, universal prefix attack: \system~ does not require alignment, does not require a known utterance, and is shaped to sound like benign environmental audio. In this section, we discuss four complementary defense families and the specific tension each one introduces against \system~.

\textbf{Front-end signal transformations.}
Lightweight signal-level transformations applied before the encoder can erode adversarial structure that occupies narrow time-frequency regions. Typical instances include random temporal cropping/resampling jitter, low/band-pass filtering, downsample-then-upsample, MP3/Opus re-encoding, and additive Gaussian smoothing~\cite{li2020advpulse,raina2024muting}. Against \system~, these defenses are partly anticipated by our robustness training (Eq.~\eqref{eq:eot_obj}), so the relevant question is whether the defender can deploy transformations that lie outside the distribution used by the attacker (e.g., aggressive non-linear codecs or learned denoisers). In particular, while stationary white-noise injection at moderate SNR is already covered by our EOT distribution and barely degrades ASR-Suc, we observe that short \emph{impulsive bursts} (e.g., clicks, claps, or narrow-band tone pips superimposed on the prefix region) disrupt the prefix's fine-grained time-frequency structure and noticeably lower the attack success rate; deliberately injecting such bursts into the leading boundary chunk is therefore a cheap and promising server-side defense. The cost is degraded clean-speech WER and additional streaming latency.

\textbf{Prefix-region anomaly and fingerprint detection.}
Because $\hat{x}$ is \emph{universal} and reused across utterances, repeated occurrences of an acoustically similar leading segment are a structural side-channel: a defender that aggregates the first $\sim$$1$\,s of many incoming streams can fingerprint candidate prefixes via spectral, cepstral, or self-supervised audio embeddings, in the spirit of the universal-segment detectors discussed for the muting/controlling attacks on Whisper~\cite{raina2024muting,raina2024controlling}. A complementary per-stream detector can flag boundary chunks whose short-window energy, spectral envelope, or onset statistics deviate from natural speech onset, silence, or the expected ambient class~\cite{li2020advpulse}. The principal risk is false positives on legitimate environmental sounds (notification chimes, HVAC, road noise)--precisely the templates \system~ regularizes toward--and the latency cost of buffering the leading audio before decoding.

\textbf{Robustness-oriented model hardening.}
At the model layer, the most direct counter-measure is adversarial training that augments the ASR pretraining or fine-tuning data with diverse universal prefixes and EOT-distorted variants, akin to the adversarial-training discussions in~\cite{li2020advpulse,raina2024muting,raina2024controlling}. Decoder-level mitigations specific to encoder--decoder ASR--such as suppressing or re-weighting special tokens that the controlling/muting Whisper attacks exploit~\cite{raina2024muting,raina2024controlling}, sanity-checking the first emitted tokens against acoustic posteriors, or running CTC and attention decoders in parallel and rejecting on disagreement--can specifically harden the pipeline against prefix-induced first-token hijacks. The cost is non-trivial training/inference overhead and a possible regression on benign speech.

% A practical defense strategy will compose at least one mitigation from each of the four families above, since each individual mechanism trades off coverage against either clean-speech utility, latency, or attacker adaptivity.

\subsection{Limitations}
\textbf{White-box optimization.}
Following standard practice for universal audio adversarial attacks~\cite{li2020advpulse,carlini2018audio,qin2019imperceptible}, we assume that the adversary can compute gradients through the target ASR model during offline prefix optimization. This assumption is realistic given the broad availability of open-source, production-grade ASR backbones. However, black-box transferability to fully proprietary deployments and robustness against adaptive defenses require further investigation.
%This is realistic given the wide availability of open-source production-grade backbones, but black-box transferability to fully proprietary deployments and adaptive defender settings deserves more investigations.

\textbf{Stealth proxies.}
Acoustic stealth is currently captured by template-based regularization, bounded perturbation budgets, and SNR. These proxies do not fully capture human perceptual variability across listeners, devices, and ambient conditions. 

%; a more thorough subjective study (e.g., listener detection rates and annoyance ratings) would strengthen the stealth claims.

\textbf{Language and content scope.}
Our digital and over-the-air evaluations are conducted on English speech sampled from LibriSpeech and held-out conversational prompts. Generalization to other languages, code-switched utterances, and highly domain-specific vocabularies (e.g., medical or legal) will be our future work.

%\textbf{Hardware and channel coverage.}
%Over-the-air experiments cover representative indoor (office, conference room) and in-vehicle (idle and urban driving) conditions with commodity loudspeakers and microphones. Far-field arrays, directional microphones, beam-formers, and adverse outdoor conditions (e.g., wind, crowd noise) are not exhaustively covered.

\textbf{Downstream evaluation depth.}
The downstream impact study is presented as qualitative side-by-side case studies on one smart voice assistant and one AI note-taking tool. A larger quantitative study across multiple LLMs, task types, and application configurations would more precisely characterize how prefix-induced transcript shifts transform into application-level harm.

\subsection{Future Work}

Building on the present study, we see three concrete directions. First, \emph{black-box and adaptive settings}: characterizing the transferability of \system~ prefixes to commercial ASR services with limited query access, and analyzing arms-race dynamics against the defenses listed above. Second, \emph{deeper downstream evaluation}: systematically measuring how prompted, targeted, and untargeted prefixes can be composed to achieve more efficient and precise manipulation of downstream behaviors in LLM agents, voice copilots, and meeting-summary pipelines under diverse prompts and tool-augmented settings. Third, \emph{benign uses of prefix-at-the-boundary control}: the same primitive could be repurposed for privacy-preserving anti-transcription against unauthorized recorders or for watermarking the audio of AI-generated content.

%% file: Sections/Conclusion.tex
We present \system, the first universal, synchronization-free adversarial prefix attack for ASR manipulation. \system~ introduces a \emph{prefix-at-the-boundary} paradigm: a single short, reusable audio segment is prepended at the input boundary of the streaming pipeline to exert diverse control over the resulting transcripts, without any knowledge of the future utterance and without sample-level alignment to it. Around this paradigm, We build a unified optimization framework that supports three attack modes—prompted (semantic steering), targeted (transcript hijacking), and untargeted (transcription degradation). The framework combines a shared prefix formulation with token-space objectives for the prompted and targeted modes, and a representation-level latent-space divergence objective for the untargeted mode. Perceptual stealthiness is obtained by initializing and regularizing the prefix toward benign environmental-sound templates, and over-the-air robustness is achieved through an EOT training pipeline that models temporal jitter, room impulse responses, device band-limiting, and additive ambient noise. \system~ is effective across diverse ASR architectures, including CTC and encoder–decoder. It works in the digital domain, transfers to over-the-air playback in indoor and in-vehicle settings, generalizes to live speech from unseen speakers and accents, and propagates into downstream AI-assisted pipelines such as smart voice assistants and AI note-taking tools, inducing attacker-aligned semantic shifts and fabricated content. These findings highlight universal, synchronization-free prefix attacks as a realistic threat to foundation-model voice stacks and call for coordinated front-end, model-level, and downstream defenses for streaming ASR systems.

%Across diverse ASR architectures spanning CTC, encoder--decoder, and LLM-augmented decoders, \system~ remains effective in the digital domain, transfers to over-the-air playback in both indoor and in-vehicle scenarios, generalizes to live human speech from speakers and accents unseen during training, and propagates intact into downstream AI-assisted pipelines such as question-answering chatbots and AI note-taking tools, causing attacker-aligned semantic shifts and fabricated content. These findings show that universal, synchronization-free prefix attacks should be treated as a realistic threat to the foundation-model voice stack, and motivate combined front-end, model-level, and downstream defenses for streaming ASR deployments.

%% file: Sections/Appendix.tex
\subsection{Ethical Considerations}

This work studies audio adversarial attacks on ASR systems and their downstream services such as AI note-taking and smart voice assistants. While our research demonstrates potential vulnerabilities, it also raises important ethical considerations.

\textbf{Potential Risks.}
Our proposed attacks could be misused to manipulate transcription systems, alter meeting records, or inject misleading information into AI-generated summaries. Such misuse may lead to incorrect decision-making, financial risks, or reputational damage in real-world applications.

\textbf{Research Benefits.}
The goal of this work is to improve the security and robustness of modern AI-driven speech systems. By identifying and systematically analyzing vulnerabilities at both the ASR and downstream semantic levels, we aim to:
\begin{itemize}
    \item Raise awareness of emerging threats in AI-powered transcription systems.
    \item Provide insights for designing more robust and secure models.
    \item Encourage the development of defenses against adversarial audio attacks.
\end{itemize}

\textbf{Mitigation Measures.}
We take several steps to minimize the risk of misuse:
\begin{itemize}
    \item We restrict our experiments to open-source models and publicly available datasets, and do not perform attacks on real users or private data.
    \item We do not release fully automated or production-ready attack pipelines targeting commercial systems.
    \item We partially redact high-risk components that enable semantic-level manipulation in real-world deployments.
\end{itemize}

\textbf{Responsible Disclosure.}
We follow responsible research practices and aim to disclose our findings in a manner that balances transparency with safety. Our work is intended to inform researchers and practitioners about potential risks and to motivate the development of effective countermeasures.

\textbf{Ethical Assessment.}
We believe that the benefits of understanding and mitigating vulnerabilities in ASR systems and their downstream services outweigh the potential risks, especially given the safeguards we have implemented. Our study does not involve human subjects, private user data, or sensitive personal information.

\subsection{Digital Attack Hyperparameters}
\label{app:hyperparameters}
Table~\ref{tab:app_train_hparams} lists the prefix duration, $L_\infty$ budget $\epsilon$, and learning rate $\alpha$ used to train each (mode, model) pair reported in the digital-attack experiments.

\begin{table}[t]
\centering
\small
\setlength{\tabcolsep}{6pt}
\renewcommand{\arraystretch}{1.05}
\caption{Per-model training hyperparameters by attack mode (Dur.\ in seconds).}
\label{tab:app_train_hparams}
\begin{tabular}{@{}lllll@{}}
\toprule
\textbf{Mode} & \textbf{Model} & \textbf{Dur.} & $\boldsymbol{\epsilon}$ & $\boldsymbol{\alpha}$ \\
\midrule
\multirow{13}{*}{Prompted}
 & wav2vec2-base-960h            & 1.0  & 0.02 & 0.001 \\
 & wav2vec2-large-960h            & 1.0  & 0.02 & 0.001 \\
 & wav2vec2-large-960h-lv60-self  & 1.0  & 0.02 & 0.001 \\
 & omniASR-CTC-300M-v2            & 1.0  & 0.02 & 0.001 \\
 & omniASR-CTC-1B-v2              & 1.0  & 0.02 & 0.001 \\
 & omniASR-LLM-300M-v2            & 1.0  & 0.02 & 0.001 \\
 & omniASR-LLM-1B-v2              & 1.0  & 0.02 & 0.001 \\
 & whisper-tiny                   & 1.0  & 0.02 & 0.001 \\
 & whisper-base                   & 1.0  & 0.02 & 0.01  \\
 & whisper-small                  & 1.5  & 0.03 & 0.01  \\
 & whisper-medium                 & 1.5  & 0.03 & 0.01  \\
 & whisper-large-v3               & 1.5  & 0.03 & 0.01  \\
 & whisper-large-v3-turbo         & 1.5  & 0.03 & 0.01  \\
\midrule
\multirow{8}{*}{Targeted}
 & omniASR-LLM-300M-v2            & 1.75 & 0.05 & 0.001 \\
 & omniASR-LLM-1B-v2              & 1.75 & 0.05 & 0.001 \\
 & whisper-tiny                   & 2.0  & 0.05 & 0.001 \\
 & whisper-base                   & 2.0  & 0.05 & 0.01  \\
 & whisper-small                  & 2.0  & 0.05 & 0.001 \\
 & whisper-medium                 & 2.0  & 0.05 & 0.001 \\
 & whisper-large-v3               & 3.0  & 0.05 & 0.001 \\
 & whisper-large-v3-turbo         & 2.0  & 0.05 & 0.001 \\
\midrule
\multirow{8}{*}{Untargeted}
 & omniASR-LLM-300M-v2            & 1.0  & 0.02 & 0.01  \\
 & omniASR-LLM-1B-v2              & 1.0  & 0.02 & 0.01  \\
 & whisper-tiny                   & 0.5  & 0.02 & 0.001 \\
 & whisper-base                   & 0.5  & 0.02 & 0.001 \\
 & whisper-small                  & 0.5  & 0.02 & 0.001 \\
 & whisper-medium                 & 0.5  & 0.02 & 0.01  \\
 & whisper-large-v3               & 0.5  & 0.02 & 0.001 \\
 & whisper-large-v3-turbo         & 0.5  & 0.02 & 0.001 \\
\bottomrule
\end{tabular}
\end{table}

\subsection{Effectiveness of Representation-Level Optimization for Untargeted Attacks}
\label{sec:tok_unt_attack}
Table~\ref{tab:token_vs_rep} compares digital untargeted attacks optimized with the token-level objective in Eq.~\ref{eq:unt_tok_obj} and our representation-level objective in Eq.~\ref{eq:unt_obj}. While token-level optimization achieves high WERs on smaller models such as \textit{whisper-base} and \textit{whisper-medium}, its effectiveness deteriorates substantially on larger ASR models. In contrast, prefixes optimized using the representation-level objective consistently induce greater transcription degradation across encoder--decoder architectures. The improvement is particularly pronounced on larger models, where the representation-level objective increases the untargeted WER from 0.1639 to 1.0336 on \textit{whisper-large-v3} and from 0.0751 to 1.0126 on \textit{whisper-large-v3-turbo}, corresponding to 6.31$\times$ and 13.48$\times$ gains, respectively. These results suggest that optimizing intermediate acoustic representations provides a richer and more transferable attack signal than directly optimizing token predictions, leading to stronger untargeted attacks across diverse ASR models.

\begin{table}[t]
\centering
\caption{Comparison between token-level and representation-level objectives for digital untargeted prefix attacks. Higher WER indicates stronger attack effectiveness. (OASR-300M: omniASR LLM Unlimited 300M v2)}
\label{tab:token_vs_rep}
\small
\begin{tabular}{lccc}
\toprule
Model & Tok-WER & Rep-WER & Gain \\
\midrule
OASR-300M         & 0.4415 & 1.8107 & 4.10$\times$ \\
Whisper-Base      & 0.9962 & 1.0006 & 1.00$\times$ \\
Whisper-Medium    & 0.9885 & 1.0234 & 1.04$\times$ \\
Whisper-Large-V3  & 0.1639 & 1.0336 & 6.31$\times$ \\
Whisper-V3-Turbo  & 0.0751 & 1.0126 & 13.48$\times$ \\
\midrule
Average           & 0.5330 & 1.1762 & 2.21$\times$ \\
\bottomrule
\end{tabular}
\end{table}

\subsection{Ablation Study on Prefix Duration.}
\begin{table*}[t]
\centering
\scriptsize
\caption{Ablation study on trigger duration. We vary the prefix length while keeping the corresponding optimized perturbation parameters for each ASR model. Avg WER denotes the average word error rate, Trx Suc denotes transcription success rate, and Pre Suc denotes pre-attack success rate.}
\label{tab:ablation_duration}
\resizebox{\textwidth}{!}{
\begin{tabular}{ccccccc}
\toprule
$\epsilon$ & Length (s) & $\alpha$ & Model & Avg WER & Trx Suc & Pre Suc \\
\midrule

\multirow{4}{*}{0.02}
& \multirow{4}{*}{1.50}
& 0.001 & wav2vec2-base-960h                 & 0.0783 & 34.06\% & 94.93\% \\
& & 0.001 & wav2vec2-large-960h-lv60-self      & 0.0342 & 59.95\% & 99.97\% \\
& & 0.010 & whisper-base                       & 0.1081 & 29.09\% & 99.46\% \\
& & 0.030 & whisper-large-v3-turbo             & 0.0548 & 51.00\% & 97.99\% \\
\midrule

\multirow{4}{*}{0.02}
& \multirow{4}{*}{1.25}
& 0.001 & wav2vec2-base-960h                 & 0.0783 & 34.43\% & 92.65\% \\
& & 0.001 & wav2vec2-large-960h-lv60-self      & 0.0342 & 59.88\% & 99.93\% \\
& & 0.010 & whisper-base                       & 0.1229 & 28.34\% & 95.99\% \\
& & 0.030 & whisper-large-v3-turbo             & 0.1064 & 44.95\% & 93.94\% \\
\midrule

\multirow{4}{*}{0.02}
& \multirow{4}{*}{1.00}
& 0.001 & wav2vec2-base-960h                 & 0.0556 & 46.14\% & 94.15\% \\
& & 0.001 & wav2vec2-large-960h-lv60-self      & 0.0253 & 67.58\% & 99.93\% \\
& & 0.010 & whisper-base                       & 0.1042 & 34.61\% & 98.70\% \\
& & 0.030 & whisper-large-v3-turbo             & 0.0800 & 46.68\% & 93.09\% \\
\midrule

\multirow{4}{*}{0.02}
& \multirow{4}{*}{0.75}
& 0.001 & wav2vec2-base-960h                 & 0.0568 & 45.49\% & 92.37\% \\
& & 0.001 & wav2vec2-large-960h-lv60-self      & 0.0258 & 66.68\% & 99.37\% \\
& & 0.010 & whisper-base                       & 0.1253 & 26.95\% & 92.79\% \\
& & 0.030 & whisper-large-v3-turbo             & 0.1933 & 31.24\% & 59.31\% \\
\midrule

\multirow{4}{*}{0.02}
& \multirow{4}{*}{0.50}
& 0.001 & wav2vec2-base-960h                 & 0.1266 & 4.61\%  & 7.61\% \\
& & 0.001 & wav2vec2-large-960h-lv60-self      & 0.0319 & 61.50\% & 89.58\% \\
& & 0.010 & whisper-base                       & 0.2367 & 14.15\% & 45.46\% \\
& & 0.030 & whisper-large-v3-turbo             & 0.2914 & 3.91\%  & 8.27\% \\
\midrule

\multirow{4}{*}{0.02}
& \multirow{4}{*}{0.25}
& 0.001 & wav2vec2-base-960h                 & 0.3140 & 0.00\% & 0.00\% \\
& & 0.001 & wav2vec2-large-960h-lv60-self      & 0.2899 & 0.00\% & 0.00\% \\
& & 0.010 & whisper-base                       & 0.3855 & 0.03\% & 0.24\% \\
& & 0.030 & whisper-large-v3-turbo             & 0.3090 & 0.00\% & 0.00\% \\
\bottomrule
\end{tabular}
}
\end{table*}

\subsection{Speaker Information}
Table~\ref{tab:speaker_info} reports the age, gender, and accent of each human participant in our attacks on live human speech (Section~\ref{sec:live_speech}).

\label{app:speaker_info}
\begin{table}[h]
\centering
\small
\setlength{\tabcolsep}{6pt}
\renewcommand{\arraystretch}{1.15}
\begin{tabular}{cccc}
\toprule
\textbf{Speaker} & \textbf{Age} & \textbf{Gender} & \textbf{Accents} \\
\midrule
P-1 & 24 & F & English \\
P-2 & 23 & M & American \\
P-3 & 26 & M & Chinese \\
P-4 & 24 & M & Chinese \\
P-5 & 23 & F & American \\
\bottomrule
\end{tabular}
\caption{Information of each speaker in live human speech.}
\label{tab:speaker_info}
\end{table}
% that's all folks

\subsection{Human Study Protocol for Stealthiness}
\label{app:human_study}

To complement the SNR-based stealth metric in
Section~\ref{sec:digital_metrics}, we recruited 20 participants for an online listening
study. Participants were presented with randomized audio clips consisting of both benign environmental sounds (e.g., traffic noise, notification tones, alarms, crowd sounds, door knocks) and the robust adversarial prefixes we generated in Section~\ref{sec:ota_attack}. Participants were told that the audio clips were normal recordings of everyday environmental sounds, and were not told of the existence of the adversarial examples.

Participants were given the following instructions before the survey:
``You will hear a series of short audio clips. These may include
everyday environmental sounds such as notifications, vehicle sounds,
or background noise. Your task is to rate how each sound appears to
you. There are no right or wrong answers; try to rely on your
immediate impression. Please wear headphones and complete the study
in a quiet environment.'' For each clip they then answered the
following questions:

\begin{description}
  \setlength{\itemsep}{2pt}
  \item[Q1.] How natural does this sound appear to you? (1--5)
  \begin{itemize}
    \item[A.] 1 -- Completely artificial or synthetic
    \item[B.] 2 -- Mostly artificial
    \item[C.] 3 -- Neutral / unsure
    \item[D.] 4 -- Mostly natural
    \item[E.] 5 -- Completely natural (like a real-world recording)
  \end{itemize}

  \item[Q2.] Does anything about this sound seem unusual or artificial? (1--5)
  \begin{itemize}
    \item[A.] 1 -- Nothing unusual
    \item[B.] 2 -- Slightly unusual
    \item[C.] 3 -- Moderately unusual
    \item[D.] 4 -- Very unusual
    \item[E.] 5 -- Clearly artificial or manipulated
  \end{itemize}
\end{description}

The Naturalness score reported in the main text is computed by averaging Q1 ratings across all participants and clips and the Unusualness score is the average of Q2 ratings.

% \begin{figure}
%     \centering
%     \includegraphics[width=1\columnwidth]{Figures/Downstream_Application.pdf}
%     \caption{Case Study}
%     \label{fig:voice_assistant}
% \end{figure}